\documentclass[review]{elsarticle}

\usepackage{lineno,hyperref}

\journal{}
\usepackage{multirow} 
\usepackage{subfigure}
\usepackage{amsfonts}
\usepackage{amsmath}
\usepackage[ruled,linesnumbered]{algorithm2e}
\usepackage{mathrsfs}
\usepackage{booktabs}
\usepackage[section]{placeins}

\begin{document}

\begin{frontmatter}

\title{FedMCSA: Personalized Federated Learning via Model Components Self-Attention}

\author{Qi Guo}
\author{Yong Qi\corref{mycorrespondingauthor}}
\cortext[mycorrespondingauthor]{Corresponding author: Yong Qi (Email: qiy@xjtu.edu.cn)}
\author{Saiyu Qi}
\author{Di Wu}
\author{Qian Li}

\address{School of Computer Science and Technology, Xi’an Jiaotong University}
\address{Xi’an, Shaanxi Province, China}

%
%
%

\begin{abstract}
Federated learning (FL) facilitates multiple clients to jointly train a machine learning model without sharing their private data. However, Non-IID data of clients presents a tough challenge for FL. Existing personalized FL approaches rely heavily on the default treatment of one complete model as a basic unit and ignore the significance of different layers on Non-IID data of clients.
In this work, we propose a new framework, federated model components self-attention (FedMCSA), to handle Non-IID data in FL, which employs model components self-attention mechanism to granularly promote cooperation between different clients. This mechanism facilitates collaboration between similar model components while reducing interference between model components with large differences. We conduct extensive experiments to demonstrate that FedMCSA outperforms the previous methods on four benchmark datasets. Furthermore, we empirically show the effectiveness of the model components self-attention mechanism, which is complementary to existing personalized FL and can significantly improve the performance of FL.
\end{abstract}


\end{frontmatter}

\section{Introduction}
Federated learning (FL) has attracted widespread attention as a paradigm of distributed learning with privacy protection~\cite{McMahan_2017_AISTATS_FedAvg,QiangYang_2019_ACMTrans_FLsurvey,TianLi_2020_SignalProcess_FLsurvey}. The standard FL follows three steps: (\romannumeral1) at each iteration, the server distributes the global model to clients;  (\romannumeral2) the client trains the local model on its local private data based on the global model; (\romannumeral3) the server aggregates local models updated by clients to achieve a new global model, repeated until convergence~\cite{McMahan_2017_AISTATS_FedAvg,AndrewHard_2018_arXiv_FL}. 
FL can ensure effective collaboration between different clients when the data distributions are independent and identically distributed (IID), i.e., private data distributions of clients are similar to each other. However, in many application scenarios, private data of clients may be different in size and class distribution, that is, the data distributions are not independent and identically distributed (Non-IID). In this case, FL may not achieve effective collaboration on different clients due to difference of individual private data~\cite{VirajKulkarni_2020_arXiv_PFL}.

\begin{figure}[t]
	\centering
	\subfigure[Average]{\label{fig.intro.1.a}
		\includegraphics[width=0.3\columnwidth]{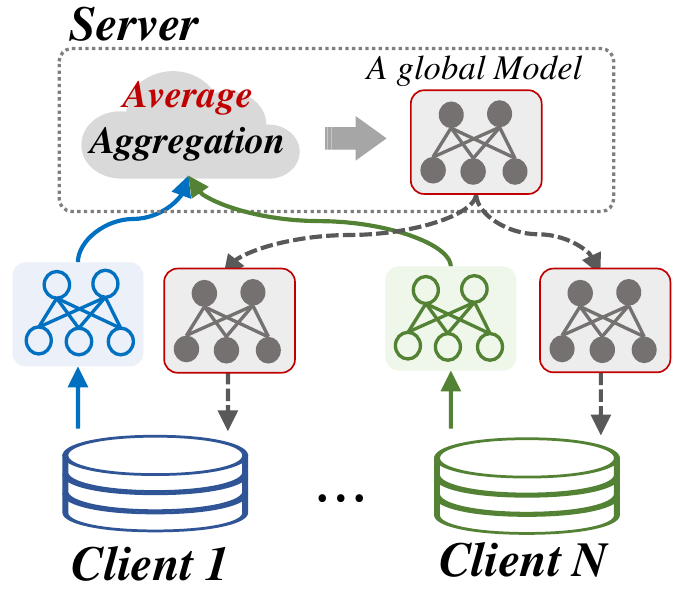}
	}
	\subfigure[Model-based]{\label{fig.intro.1.b}
		\includegraphics[width=0.3\columnwidth]{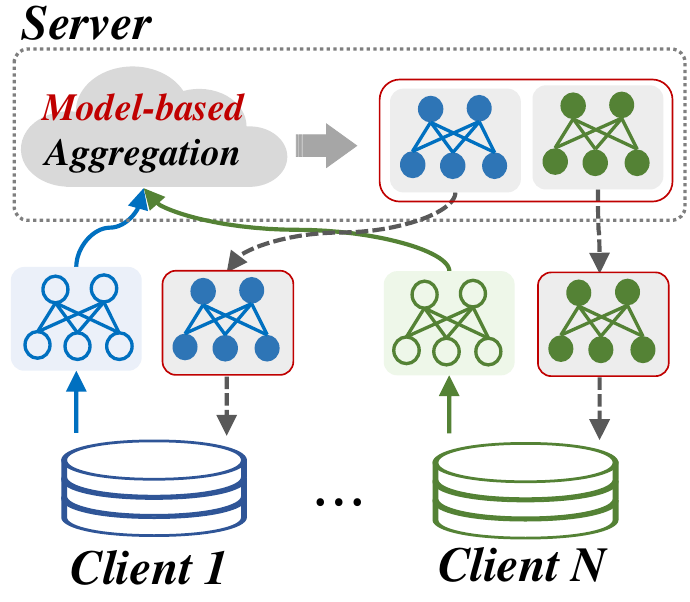}
	}
	\subfigure[Component-based (Our)]{\label{fig.intro.1.c}
		\includegraphics[width=0.3\columnwidth]{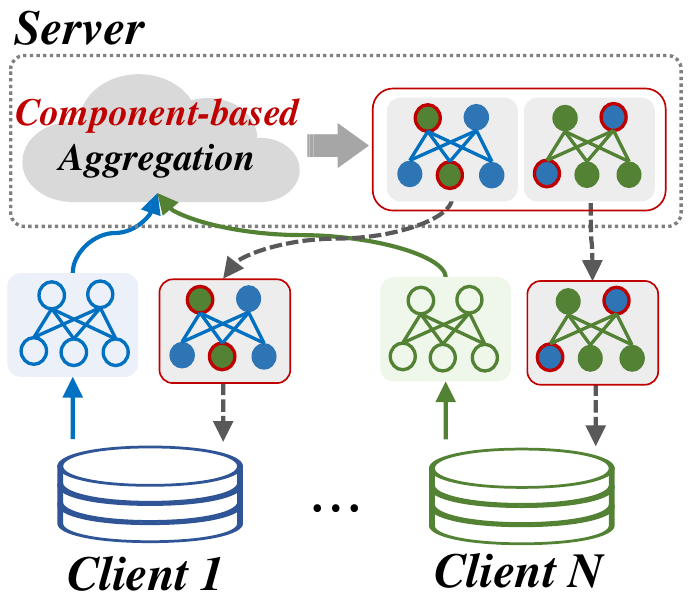}
	}

	\caption{The illustrations of different FL methods.}
	\label{fig.intro.1}
\end{figure}

Various algorithms have been proposed to handle the Non-IID data in FL, which can be divided into two categories: average aggregation methods and model-based aggregation methods. 
As shown in \figurename~\ref{fig.intro.1.a}, average aggregation methods average all local models to generate a global model and distribute it to all clients, where an additional fine-tuning step is performed to train the personalized model in the clients~\cite{wang2019federated,mansour2020three,AlirezaFallah_2020_NIPS_PFL, CanhTDinh_2020_NIPS_PFL}. However, using one global model is difficult to fit different clients with Non-IID data. As a result, as illustrated in~\figurename~\ref{fig.intro.1.b}, model-based aggregation methods weight different local models to generate a personalized global model for each client, which treats the entire model as a basic unit to calculate the weighting coefficient of the local model~\cite{MichaelZhang_2021_ICLR_PFL, YutaoHuang_2021_AAAI_PFL, FilipHanzely_2020_NIPS_PFL}. Nevertheless, these methods ignore significance of different layers within the model and cause the curse of dimensionality when computing the similarity of high-dimensional models~\cite{verleysen2005curse}.

We argue each layer’s significance should be considered for handling Non-IID data of clients. It is inappropriate to disregard significance of layers and treat all layers equally by model-based aggregation methods on the server.
As illustrated in \figurename~\ref{fig.intro.1.c}, we regard each layer in a model as a basic unit, i.e., a model component, and present a component-based aggregation method to granularly facilitate collaboration between different clients. 

In this work, we propose a novel framework, \emph{federated model components self-attention} (FedMCSA), to handle Non-IID data in FL. 
The core of FedMCSA is a model components self-attention mechanism, which utilizes the component-based aggregation method to adaptively update model components on the server. 
This mechanism facilitates collaboration between similar model components while reducing interference between model components with large differences.
Specifically, FedMCSA first decomposes the local models from clients to obtain model components on the server, then lets the model components perform parallel self-attention operations, and finally, generates complete personalized models to send them to the clients. In this way, FedMCSA achieves a complete personalized FL and promotes purposeful and efficient collaboration among clients. The experimental results not only show that FedMCSA outperforms FedAvg~\cite{McMahan_2017_AISTATS_FedAvg}, Fedprox~\cite{TianLi_2020_Mlsys_FL}, Per-FedAvg~\cite{AlirezaFallah_2020_NIPS_PFL}, pFedMe~\cite{CanhTDinh_2020_NIPS_PFL}, and HeurFedAMP~\cite{YutaoHuang_2021_AAAI_PFL} in different settings, but also empirically demonstrate the effectiveness of the model components self-attention mechanism.

Our contributions and novelty can be summarized as follows:
\begin{itemize}
	\item We propose a novel framework, federated model components self-attention (FedMCSA), to handle Non-IID data in FL, which can achieve a complete personalized FL to adaptively update models.
	
	\item We devise a new model components self-attention mechanism to granularly address Non-IID data from the perspective of the internal layers of the model, which can be seamlessly integrated into FL.
	
	\item Extensive experiments on four datasets are conducted
	to compare the proposed FedMCSA with state-of-the-art
	methods as well as its ablation variants. The results
	suggest that FedMCSA achieves a significant improvement in performance for personalized FL.
\end{itemize}

\section{Related Works}
\textbf{FL}
The first FL algorithm is FedAvg~\cite{McMahan_2017_AISTATS_FedAvg}, which is an iterative algorithm of client-server architecture. The current techniques aim to train a global model through cooperation between clients without leaking their private data to other clients, which can achieve better performance than working alone. Various challenges in FL have been investigated and addressed, including privacy protection~\cite{LigengZhu_2019_NIPS_attack,JinhyunSo_2020_NIPS_FLPrivacy,StaceyTruex_2019_AISec_FLPrivacy, AlekseiTriastcyn_2019_BigData_FLPrivacy} and communication complexity~\cite{JennyHamer_2020_ICML_FLCommunication,DanielRothchild_2020_ICML_FLCommunication,GrigoryMalinovskiy_2020_ICML_FLCommunication}. 
A main challenge of FL is statistical diversity, which means that data distributions among clients are Non-IID with affecting its performance and convergence rate~\cite{AlirezaFallah_2020_NIPS_PFL,RuiyuanWu_2021_AAAI_PFL,JinzeWu_2021_WWW_PFL,pmlr-v139-collins21a,pmlr-v139-shamsian21a}.

\textbf{Personalized FL}
Diverse methods have been proposed to address the problem of personalized FL. Fedprox~\cite{TianLi_2020_Mlsys_FL} adds a proximal term to the objective to address the challenges of heterogeneity. The goal of Per-FedAvg~\cite{AlirezaFallah_2020_NIPS_PFL} is to get a global model as initialization, and then one more step of gradient update in each client is performed to obtain the personalized model. pFedMe~\cite{CanhTDinh_2020_NIPS_PFL} uses the Moreau envelope as the clients’ regularized loss function to achieve the decoupling of personalized model optimization and global model learning, which formulates a bi-level optimization problem in the client for personalized FL. The method for training a mixture of local and global models is considered as a personalized solution for each client~\cite{FilipHanzely_2020_arXiv_FLMixture,YuyangDeng_2020_arXiv_PFL}. Under the assumption that the private model of a client is provided to other clients in addition to the server, the optimal weight combination based on the mutual benefits between the models is calculated for each client to achieve personalization~\cite{MichaelZhang_2021_ICLR_PFL}, which can work better than computing a single model average with constant weights for the entire federation as in traditional FL. For the cross-silo scenario, FedAMP and HeurFedAMP~\cite{YutaoHuang_2021_AAAI_PFL} design a federated attentive message passing method to conduct personalized FL with preserving privacy. However, FedAMP and HeurFedAMP both need to understand the clients’ data distribution to set constant weight hyperparameters, which is difficult in actual applications. Meta-learning can also be used for personalization~\cite{AlirezaFallah_2020_NIPS_PFL, SenLin_2020_ICDCS_FLMetaLearning}. The objective of the work ~\cite{AlirezaFallah_2020_NIPS_PFL} is to investigate a personalized variant of FL, whose purpose is to develop an initial shared model that existing or prospective clients can easily modify to fit their local datasets by performing one or several steps of gradient descent on their own data. The authors propose a federated multi-task framework called MOCHA~\cite{VirginiaSmith_2017_NIPS_FLMTL} using multi-task learning to address data heterogeneity and communication efficiency. For more details about FL and personalized FL, we recommend referring to these comprehensive surveys~\cite{TianLi_2020_SignalProcess_FLsurvey,PeterKairouz_2019_arXiv_FLProblems}, which not only investigate the problem of heterogeneity for personalized FL, but also discuss the unique characteristics and challenges of FL, provide a broad overview of current approaches, and present an extensive collection of future work.

\textbf{Self-Attention}
The attention mechanism is widely used in many fields, which fully demonstrates the effectiveness of attention~\cite{AshishVaswani_2017_NIPS_attention, AlanadeSantanaCorreia_2021_arXiv_attentionsurvey}. The self-attention mechanism is a variant of the attention mechanism, which reduces the dependence on external information and is better at capturing the internal correlation of data or features~\cite{DavidWRomero_2021_ICLR_attentionvision, YaruHao_2021_AAAI_selfattention}. 

Different from existing works, we have deeply explored the potential of the interaction between personalized models from the perspective of the internal components of the model, and then achieve the adaptive update of different personalized models by using our specially designed FedMCSA.

\section{Problem Definition of Personalized Federated Learning}
The conventional formulation of FL aims to find a global model $\Theta$ by minimizing the overall population loss
\begin{equation}
	\min _{\Theta \in \mathbb{R}^{d}}\left\{f(\Theta):=\frac{1}{N} \sum_{i=1}^{N} f_{i}(\Theta)\right\}, \label{eq:problem.new.1}
\end{equation}
where the function $f_{i}: \mathbb{R}^{d} \rightarrow \mathbb{R}, i=1, \ldots, N$, denotes the expected loss of the client $i$ that only depends on her/his own data distribution of ${\xi _i}$. 

Instead of solving the conventional FL problem (\ref{eq:problem.new.1}), we aim to adaptively solve personalized FL with the Non-IID private data from the perspective of internal model components. Motivated by~\cite{MichaelZhang_2021_ICLR_PFL,YutaoHuang_2021_AAAI_PFL,FilipHanzely_2020_NIPS_PFL}, we allow each client to own a personalized model on the server, which does not depend on the single global model. Consider that $N$ clients $C_1$, ..., $C_N$ corresponding to $N$ datasets $D_1$, ..., $D_N$ have their own personalized models ${\Theta _1}$, ..., ${\Theta _N}$ under the same model structure. For a neural network typically consisting of $L$ layers, the model parameter of $l$-th layer of ${\Theta _i}$ is denoted as ${\theta _{i,l}}$. The purpose of the personalized FL problem is to use the private training data and the component-based aggregation method to train the personalized model ${\Theta _i}$ to make it close to the best performance that can be achieved on the distribution of ${\xi _i}$. We want to allow similar model components collaborate more and reduce interference between model components with large differences.
Accordingly, each personalized model could learn useful knowledge from other models as much as possible to improve the performance of FL. Therefore, the personalized FL problem is formulated as
\begin{equation}
	\mathop {\min }\limits_{U  = [{\Theta _1},...,{\Theta _N}]}  \left\{ {F(U ): = \underbrace {\frac{1}{N}\sum\limits_{i = 1}^N {{f_i}} \left( {{\Theta _i}} \right)}_{: = f(U )} + \gamma \underbrace {\frac{1}{{2N}}\sum\limits_{l = 1}^L {\sum\limits_{i \ne j}^N d \left( {{{\left\| {{\theta _{i,l}} - {\theta _{j,l}}} \right\|}^2}} \right)} }_{: = \phi (U )}} \right\}, \label{eq:problem.new.2}
\end{equation}
where ${U  = [{\Theta _1},...,{\Theta _N}]}$ is a personalized model set, $f_{i}$ is the expected loss of the client $i$ corresponding to the data distribution ${\xi _i}$,  $\gamma$ is a regularization parameter, and $d\left( {{{\left\| {{\theta _{i,l}} - {\theta _{j,l}}} \right\|}^2}} \right)$ is the difference measurement function between model component ${\theta _{i,l}}$ and ${\theta _{j,l}}$. According to the work~\cite{YutaoHuang_2021_AAAI_PFL}, here we naturally assume that $d:[0,\infty ) \to \mathbb{R}$ is a non-linear function with the properties that $d$ is increasing and concave on $[0,\infty )$, continuously differentiable on $(0,\infty )$, and  $\lim _{t \rightarrow 0^{+}} d^{\prime}(t)$ is finite by $d(0)=0$. 

As to be illustrated in the next section, our novel use of model components self-attention mechanism facilitates adaptively collaboration between clients by promoting similar model components to collaborate more with each other. Furthermore, the model components self-attention mechanism is agnostic to the clients' private data distribution, which allows each client to perform on arbitrary target distributions according to requirements. The model components self-attention mechanism not only boosts the performance of personalized FL dramatically but also can be easily integrated to further improve the performance of existing personalized FL with average aggregation method. 

\section{Methodology}
In this section, to tackle the optimization problem in (\ref{eq:problem.new.2}), we first give a general method without considering privacy preservation. Then, we implement the process of the general method by proposing a new personalized FL framework, federated model components self-attention (FedMCSA), which adaptively conducts personalized FL from the perspective of the internal components of the model using parallel self-attention while protecting clients' private data. 
\subsection{The General Method}
Considering that $f(U ) = \frac{1}{N}\sum\limits_{i = 1}^N {{f_i}} \left( {{\Theta _i}} \right)$ and $\phi (U ) = \frac{1}{{2N}}\sum\limits_{l = 1}^L {\sum\limits_{i \ne j}^N d \left( {{{\left\| {{\theta _{i,l}} - {\theta _{j,l}}} \right\|}^2}} \right)} $, we can rewrite the optimization problem in (\ref{eq:problem.new.2}) to 
\begin{equation}
	\mathop {\min }\limits_U  \left\{ {F(U ): = f(U ) + \gamma \phi (U )} \right\}. \label{eq:problem.new.3}
\end{equation} 
A natural way to tackle the optimization problem in (\ref{eq:problem.new.3})  is to adopt the framework of incremental-type optimization~\cite{DimitriPBertsekas_2015_arXiv_Optimizationincrementaltype,YutaoHuang_2021_AAAI_PFL}, which can be realized by iteratively optimize $F(U )$ by alternatively optimizing $f(U )$ and $\phi (U )$ until convergence. Specifically, the general method first optimize ${\phi (U )}$ by applying a gradient descent step, and then optimize ${f(U )}$ by applying a proximal point step.
In the $t$-th iteration, the update is formulated as
\begin{equation}
	{V^t} = {U ^{t - 1}} - {\tau _t}\nabla \phi ({U ^{t - 1}}), \label{eq:problem.new.4}
\end{equation}
\begin{equation}
	{U ^t} = \arg \mathop {\min }\limits_U  \left\{ {f(U ) + \frac{\gamma }{{2{\tau _t}}}{{\left\| {U  - {V^t}} \right\|}^2}} \right\}, \label{eq:problem.new.5}
\end{equation}
where ${V^t}$ denotes the prox-center, and ${\tau _t} > 0$ is the step size of gradient descent. 
The general method has been proven in prior work ~\cite{YutaoHuang_2021_AAAI_PFL} to converge to the optimal solution when $F(U )$ is convex and to a stationary point when $F(U )$ is nonconvex.
\subsection{Federated Model Components Self-Attention}
\begin{figure*}[t]
	\centering
	\includegraphics[width=0.99\columnwidth]{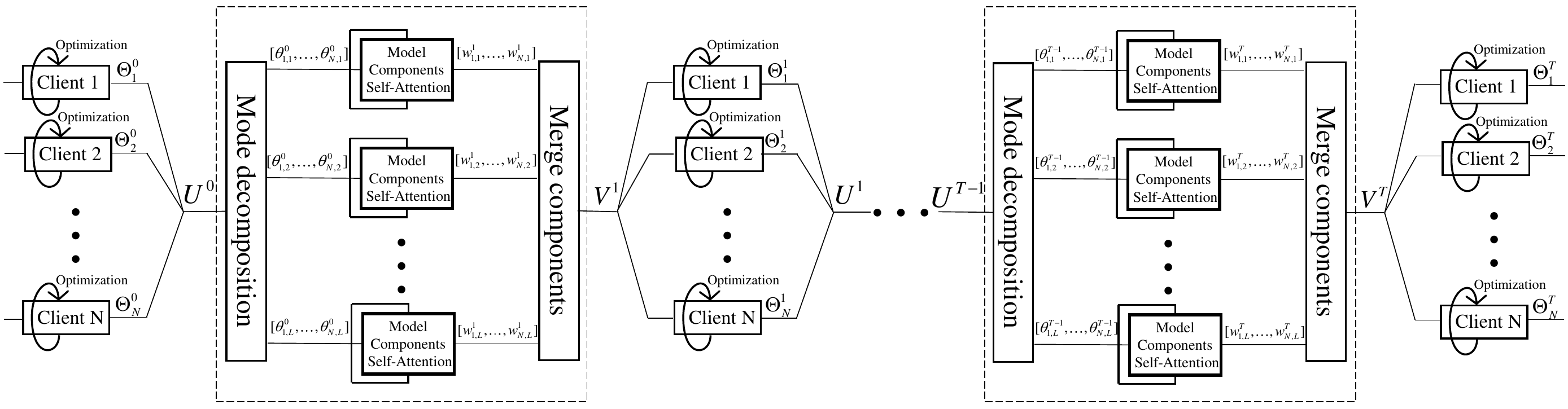}
	\caption{The workflow of federated model components self-attention for personalized federated learning. (At each iteration, the clients perform gradient descent based on the distributed personalized models to generate the updated models, which are then sent to the server. After receiving latest models from clients, the server update personalized models by the model components self-attention mechanism, and distributes the new personalized models to the corresponding clients.)}
	\label{fig.alg.1}
\end{figure*}

The general method can be easy to deploy by gathering all clinets' data together. In this subsection, we propose FedMCSA to optimize the personalized FL problem while protecting the data privacy of each client. Specifically, FedMCSA maintains a personalized cloud model for each client on a cloud server to implement the optimization step (\ref{eq:problem.new.4}) of the general method and deploys the optimization step (\ref{eq:problem.new.5}) privately in each client. The workflow of FedMCSA is illustrated in \figurename~\ref{fig.alg.1}.

Following the optimization steps of the general method, FedMCSA firstly aims to optimizes ${\phi (U )}$ and implements the optimization step in (\ref{eq:problem.new.4}) by computing matrix ${V^t}$ on the cloud server. Let ${V^t} = [W_1^t,...,W_N^t]$ and $W_i^t=[w_{i,1}^t \parallel w_{i,2}^t \parallel ... \parallel w_{i,L}^t]$, where $W_1^t,W_2^t,...,W_N^t$ are the columns of ${V^t}$ and $w_{i,1}^t, w_{i,2}^t,..., w_{i,L}^t$ are the layers of $W_i^t$, the update of the $i$-th column and $l$-th layer $w_{i,l}^t$ of ${V^t}$ computed in (\ref{eq:problem.new.4}) can be rewritten into a linear combination of the model parameter sets ${\theta _{1,l}^{t-1}}, \ldots ,{\theta _{N,l}^{t-1}}$ as follows
\begin{equation}
	\begin{aligned}
	w_{i,l}^t =& \left( {1 - \frac{{{\tau _t}}}{N}\sum\limits_{j \ne i}^N {{d^\prime }} \left( {{{\left\| {\theta _{i,l}^{t - 1} - \theta _{j,l}^{t - 1}} \right\|}^2}} \right)} \right) \cdot \theta _{i,l}^{t - 1} + \frac{{{\tau _t}}}{N}\sum\limits_{j \ne i}^N {{d^\prime }} \left( {{{\left\| {\theta _{i,l}^{t - 1} - \theta _{j,l}^{t - 1}} \right\|}^2}} \right) \cdot \theta _{j,l}^{t - 1} \\
	=& \psi _{i,l,1}^t\theta _{1,l}^{t - 1} +  \cdots  + \psi _{i,l,N}^t\theta _{N,l}^{t - 1},
\end{aligned}
 \label{eq:problem.new.x}
\end{equation}
where $d^{\prime}\left(\left\|\theta_{i,l}^{t-1}-\theta_{j,l}^{t-1}\right\|^{2}\right)$ is the derivative of $d\left(\left\|\theta_{i,l}^{t-1}-\theta_{j,l}^{t-1}\right\|^{2}\right)$ and ${\psi _{i,l,1}^{t}}, \ldots ,{\psi _{i,l,N}^{t}}$ are the linear combination weights of the model parameter ${\theta _{1,l}^{t-1}}, \ldots ,{\theta _{N,l}^{t-1}}$, respectively. Since ${\psi _{i,l,1}^{t}} +  \ldots  + {\psi _{i,l,N}^{t}} = 1$, $w_{i,l}^{t}$ is actually a convex combination~\cite{YutaoHuang_2021_AAAI_PFL} of the model parameter ${\theta _{1,l}^{t-1}}, \ldots ,{\theta _{N,l}^{t-1}}$.

We have argued for finding a linear combination of the model components to update the model component. The main question that follows is how to compute the effective weight coefficients in such a way that FedMCSA can seamlessly be incorporated into any existing FL paradigm. It is vital to determine how to calculate each model component's weight coefficient as it has a crucial effect on the output of each model component and the performance of personalized FL. 

A native method is to refer to HeurFedAMP to fix the weight coefficient for the model component that needs to be updated currently, set it as a hyperparameter, and then assign the remaining weights to other model components according to the similarity between the current model component and other model components. The setting of this hyperparameter needs to consider the clients' data distribution, for example, it is set to $1/(M_i+1)$, where $M_i$ is the number of same distribution clients for client $i$. Unfortunately, in practice, it is often difficult to provide this knowledge about the clients' data distribution. Moreover, in the actual model components update process, the interaction between different model components is dynamic, therefore fixing the weight coefficient limits the dynamic change in the weight coefficients, which artificially damages the performance of the personalized models. 

To avoid the requirement for clients' data distribution and inappropriate restrictions on the weight coefficients of model components, FedMCSA introduces self-attention into the update of model components, which performs $L$ self-attention processes in parallel to produce the updated personalized model for each client. FedMCSA makes no assumptions about the knowledge of underlying data distributions or client similarities to provide greater flexibility in personalization. Specifically, for each self-attention process, $N$ model components $[{\theta _{1,l}^{t-1}}, \ldots ,{\theta _{N,l}^{t-1}}]$ corresponding to $l$-th layer from clients are used as input, and the output is the updated $N$ model components $[{w _{1,l}^{t}}, \ldots ,{w _{N,l}^{t}}]$ for $l$-th layer. The $query$ vector, $key$ vector, and $value$ vector are denoted by $[{\theta _{1,l}^{t-1}}, \ldots ,{\theta _{N,l}^{t-1}}]$ for $l$-th layer in each model components self-attention process. 

Meanwhile, FedMCSA adopts the simple but very efficient cosine similarity function $sim(query, key) = \sigma \cos (query, key)$ with a scale hyperparameter $\sigma$ as the metric function between different model components. Therefore, ${\psi _{i,l,1}^{t}},  \ldots, {\psi _{i,l,N}^{t}}$ in (\ref{eq:problem.new.x}) can be expressed as the following

\begin{equation}
\psi _{i,l,k}^t = \frac{{{e^{\sigma \cos (\theta _{i,l}^{t - 1},\theta _{k,l}^{t - 1})}}}}{{\sum\nolimits_{h = 1}^N {{e^{\sigma \cos (\theta _{k,l}^{t - 1},\theta _{h,l}^{t - 1})}}} }},k \in [1,N].
\end{equation}

For simplicity, we denote the operation of FedMCSA on the server as the function $\mathscr{F}_{M\!C\!S\!A}$, where $U^{t-1}$ and $V^{t}$ are the input and output of $\mathscr{F}_{M\!C\!S\!A}$, respectively. Therefore, (\ref{eq:problem.new.4}) can be reformulated as the following
\begin{equation}
	V^{t}=\mathscr{F}_{M\!C\!S\!A}\left(U^{t-1}\right). \label{eq:problem.new.7} 
\end{equation}

\begin{algorithm}[htb]
	\caption{FedMCSA}\label{algorithm.new.1}
	\KwIn{$N$ clients with private training data, $T$, $R$, $S$, $\lambda$, $\eta$, $\sigma$, $U^{0}= [\Theta _1^0, \ldots ,\Theta _N^0]$}
	\KwOut{The personalized models ${U ^T} = [\Theta _1^T, \ldots ,\Theta _N^T]$}
	\For{$t = 1$ to $T$}{
		\tcp{Server}
		Server uniformly samples a subset of clients $\mathcal{S}^{t}$ with size $S$, and each of the sampled client sends the local model ${\Theta}_{i}^{t-1}, \forall i \in \mathcal{S}^{t}$, to the server\;
		
		Server using the model components self-attention mechanism computes ${V^t} = \mathscr{F}_{M\!C\!S\!A}({U ^{t - 1}})$ in (\ref{eq:problem.new.7}) by ${U ^{t-1}} = \{{\Theta}_{i}^{t-1}, \forall i \in \mathcal{S}^{t}\}$\;
		Server sends ${W}_{i}^{t}, \forall i \in \mathcal{S}^{t}$ to the clients, respectively\;
		
		\tcp{Client}
		\For{all $i = 1$ to $N$}{
			\If{The client receives the updated personalized model sent by the server}{
				$\Theta_{i}= W_i^t$\;
				$W_{l\!o\!c\!a\!l}^t = W_i^t$\;
			}
			\For{$r = 0$ to $R - 1$}{
				Sample a fresh mini-batch $\mathcal{D}_{i}$ with size $|\mathcal{D}|$ to compute $\Theta _i^{t,r} = \arg \mathop {\min }\limits_{\Theta _i  \in \mathbb{R}{^d}} \left\{ {{f_i}({\Theta _i}) + \frac{\lambda }{2}{{\left\| {{\Theta _i} - W_{l\!o\!c\!a\!l}^t} \right\|}^2}} \right\}$ defined in (\ref{eq:problem.new.8})\;
				
				$\Theta_i = \Theta _i^{t,r}$\;
				
			}
			$\Theta_i^t = \Theta _i^{t,r}$.
		}
		
	}
\end{algorithm}
After optimizing $\phi (U )$ using the model components self-attention mechanism on the server, we further implement the optimization of $f (U )$ in the client.
Let ${U^t} = [\Theta_1^t,...,\Theta_N^t]$, where $\Theta_1^t,...,\Theta_N^t$ are the columns of ${U^t}$, the update of the $i$-th column $\Theta_{i}^t$ of ${U^t}$ computed in (\ref{eq:problem.new.5}) can be rewritten as the following
\begin{equation}
	\Theta _i^t = \arg \mathop {\min }\limits_{\Theta _i  \in \mathbb{R}{^d}} \left\{ {{f_i}({\Theta _i}) + \frac{\lambda }{2}{{\left\| {{\Theta _i} - W_i^t} \right\|}^2}} \right\}, \label{eq:problem.new.8}
\end{equation}

where $\lambda  = \frac{\gamma }{{{\tau _t}}}$. We optimize $\phi (U )$ by (\ref{eq:problem.new.7}) on the server and $f (U )$ by (\ref{eq:problem.new.8}) in the client, which together constitute the complete personalized FL framework FedMCSA. In the continuous iterative process of personalized FL, the entire process can be considered as a continuous federated model components self-attention network, as shown in \figurename~\ref{fig.alg.1}.

The complete algorithm description of FedMCSA is presented in Algorithm \ref{algorithm.new.1}. FedMCSA implements a client-server personalized FL framework to boost the performance of personalized FL dramatically, which alternately optimizes $\phi (U )$ and $f (U )$ through the model components self-attention mechanism on the server and the proximal gradient descent method in the client until a maximum number $T$ of iterations is reached.

\section{Experimental Results and Discussion}
In this section, we evaluate the performance of FedMCSA and demonstrate the effectiveness of the model components self-attention mechanism when the distributions of clients' private data are Non-IID. First, we compare FedMCSA with the existing personalized FL algorithms under various datasets with different network models. Then we demonstrate the the effectiveness by ablation and integration studies. Finally, the impact of important hyperparameter $\sigma$ is analyzed on the convergence of FedMCSA.
\subsection{Experimental Settings}
\subsubsection{Datasets}
We use four public benchmark datasets that are widely used in FL. 

\begin{itemize}
	\item Synthetic~\cite{CanhTDinh_2020_NIPS_PFL}: The Synthetic dataset is applied to a 10-class classifier, where each data point is composed of 60-dimensional real-valued data. The data size of each client is in the range of [250, 25810]. The data generation and distribution procedure is adopted from the previous works~\cite{CanhTDinh_2020_NIPS_PFL,TianLi_2020_Mlsys_FL} to generate Non-IID data, which uses two parameters $\overline \alpha$ and $\overline \beta$ to control the difference of the dataset and the local model of each client. Specifically, a synthetic dataset with $\overline \alpha = 0.5$ and $\overline \beta = 0.5$ is generated and distributed according to the power law to $N = 100$ clients~\cite{CanhTDinh_2020_NIPS_PFL}. 
	
	\item Mnist~\cite{lecun1998gradient}: The Mnist dataset that is a handwritten digit dataset contains 70,000 instances with 10 labels. We adopt the Non-IID setup and data generation procedure in the work~\cite{CanhTDinh_2020_NIPS_PFL}, where the complete dataset is distributed to $N = 20$ clients owing to the limitation on MNIST’s data size. Each client's data size is different in the range of [1165, 3834] with only 2 classes of the 10 classes.
	
	\item FMnist (Fashion-Mnist)~\cite{xiao2017fashion}: The FMNIST dataset includes 70,000 fashion products across ten categories, with 7,000 images per category. We use a same generation method as the Mnist dataset to produce the Non-IID data of FMnist.
	
	\item Cifar10~\cite{krizhevsky2009learning}: The Cifar10 dataset contains 60000 32x32 color images, each of which is categorized into one of ten mutually exclusive labels. We use a same generation method as the Mnist dataset to produce the Non-IID data of Cifar10.
\end{itemize}

\subsubsection{Network Models}
To demonstrate the generality and effectiveness of our proposed FedMCSA, we consider two different models simultaneously following the work~\cite{CanhTDinh_2020_NIPS_PFL} in the experiments. First, a ${l_2}$-regularized multinomial logistic regression model (MLR) is implemented with the softmax activation and cross-entropy loss functions. Meanwhile, we consider a two-layer deep neural network (DNN) with a hidden layer of size 20 for Synthetic and 100 for MNIST, FMnist, and Cifar10 using ReLU activation and a softmax layer at the end. 
\subsubsection{Baselines}
We compare FedMCSA against five methods broadly falling under three categories. (\romannumeral1) Only train a single global model for all clients: FedAvg~\cite{McMahan_2017_AISTATS_FedAvg} and Fedprox~\cite{TianLi_2020_Mlsys_FL}. (\romannumeral2) Train more than one model but based on a global model: Per-FedAvg~\cite{AlirezaFallah_2020_NIPS_PFL} and pFedMe~\cite{CanhTDinh_2020_NIPS_PFL}. (\romannumeral3) Train a separate model for each client without considering a single global model: HeurFedAMP~\cite{YutaoHuang_2021_AAAI_PFL}. The details of the baselines are presented as follows.
\begin{itemize}
	\item FedAvg~\cite{McMahan_2017_AISTATS_FedAvg}: As the most popular FL baseline, it directly averages all local models on the server to obtain a new global model.
	
	\item Fedprox~\cite{TianLi_2020_Mlsys_FL}: A proximal term is added to the objective to address the challenges of the Non-IID data.
	
	\item Per-FedAvg~\cite{AlirezaFallah_2020_NIPS_PFL}: After getting a global model as initialization, each client performs one more step of gradient update to obtain the personalized model.
	
	\item pFedMe~\cite{CanhTDinh_2020_NIPS_PFL}: The Moreau envelope is used as the clients’ regularized loss function to achieve the decoupling of personalized model optimization and global model learning. For the comparison with pFedMe, we use both its personalized model pFedMe(PM) and global model pFedMe(GM).
	
	\item HeurFedAMP~\cite{YutaoHuang_2021_AAAI_PFL}: A federated attentive message passing method is designed to conduct personalized FL with preserving privacy.
\end{itemize}

\subsubsection{Implementation Details}
We utilize the SGD optimizer and use PyTorch to implement our method. We randomly split all datasets into 75$\%$ and 25$\%$ for training and testing, respectively. The learning rates are set to $\eta$ = 0.02 by default for all four benchmark datasetrs. Meanwhile, the batch size is set to $|\mathcal{D}|$ = 20, and the local training epoch is set to $R$ = 20. The value of hyperparameter $\sigma$ is set to the default value of 50, and the value of hyperparameter $\lambda$ is set to the default value of 5. The subset of all clients is set to $S=20$ for Synthetic, $S=10$ for Mnist, FMnist, and Cifar10. The maximum number $T$ of iterations is set to 800 for all four benchmark datasetrs.  The performance of all the methods is evaluated by considering the best mean testing accuracy (BMTA) in percentages, where BMTA is the highest mean testing accuracy achieved by a method during all communication rounds of training, and the mean testing accuracy is defined as the average of the testing accuracy on all clients. All experiments are implemented in PyTorch 1.7 running on Intel(R) Xeon(R) CPU, 64G memory, NVIDIA 1080Ti, and Ubuntu 16.04.

\subsection{Performance Comparison}
To demonstrate the empirical superiority of FedMCSA, several comparisons between FedMCSA, HeurFedAMP, pFedMe, Per-FedAvg, Fedprox, and FedAvg are conducted. First of all, the same hyperparameters for all algorithms are used as a basic comparison. Considering that the performance of algorithms will behave differently when the hyperparameters change, we further perform a grid search of hyperparameters to find the highest performance of combination fine-tuning hyperparameters.
\subsubsection{The comparisons for the same hyperparameters.}
\begin{figure}[t]
	\centering
	\subfigure[]{
		\includegraphics[width=0.35\columnwidth]{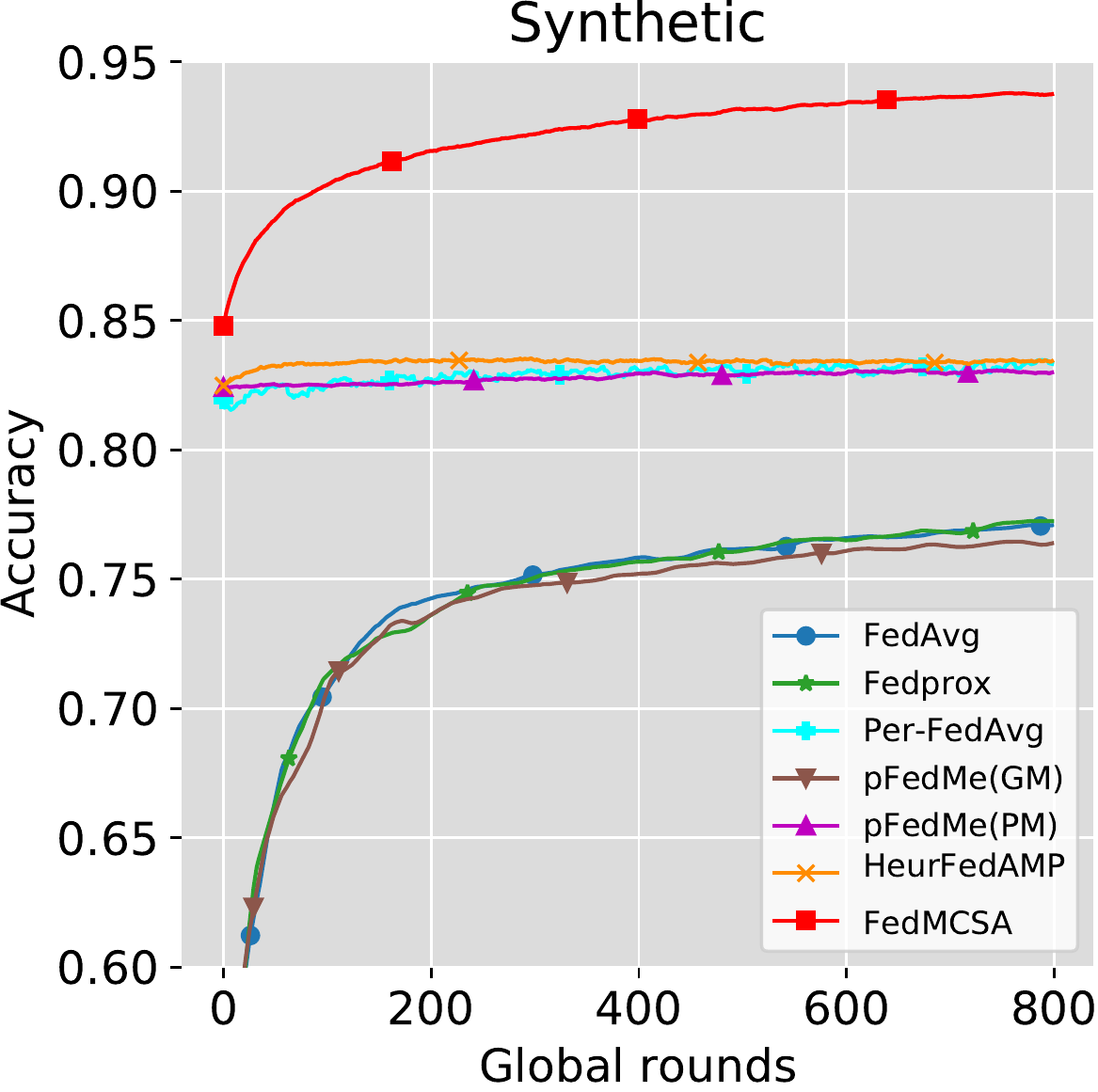}
	}
	\subfigure[]{
		\includegraphics[width=0.35\columnwidth]{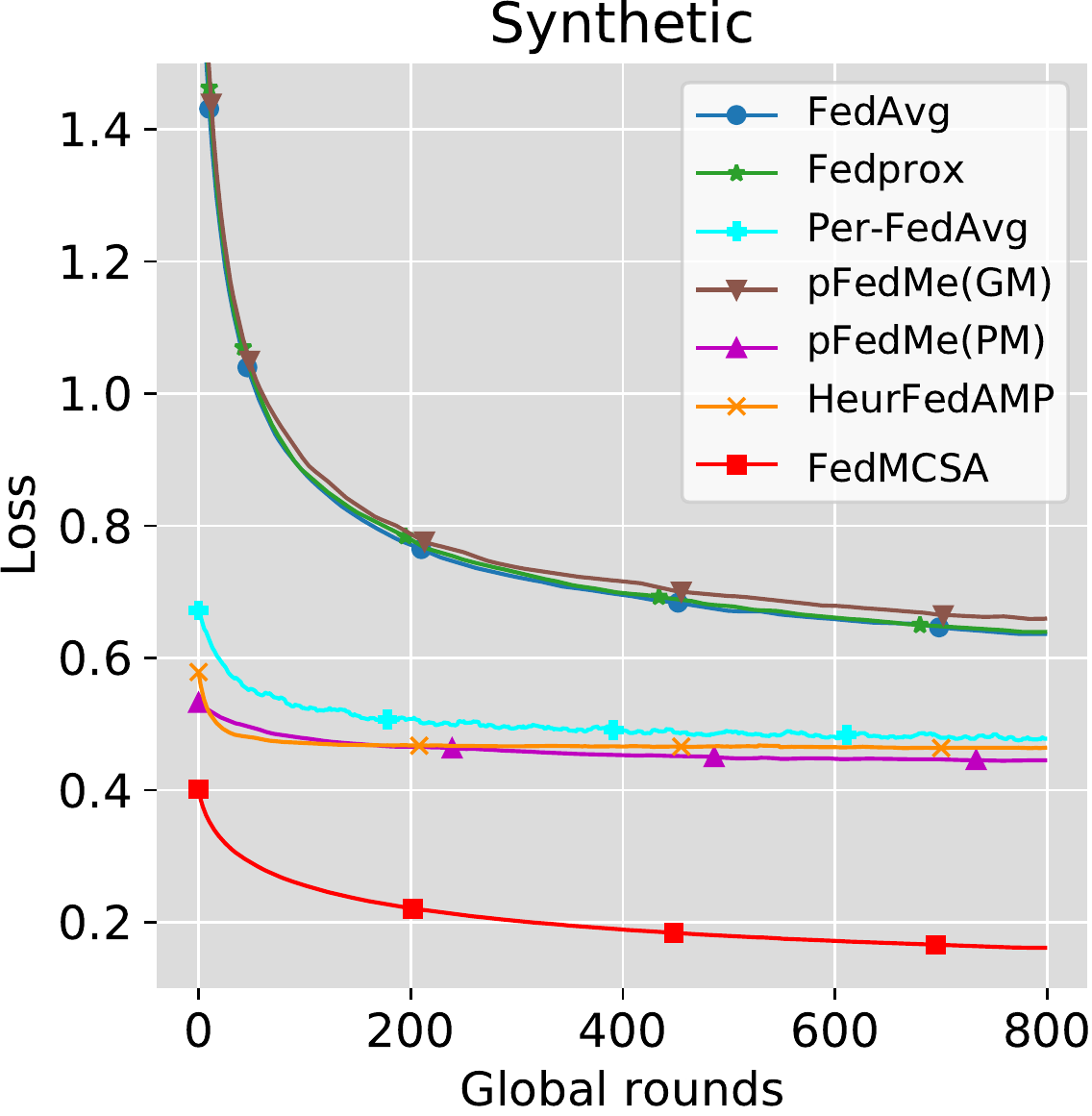}
	}
	\subfigure[]{
		\includegraphics[width=0.35\columnwidth]{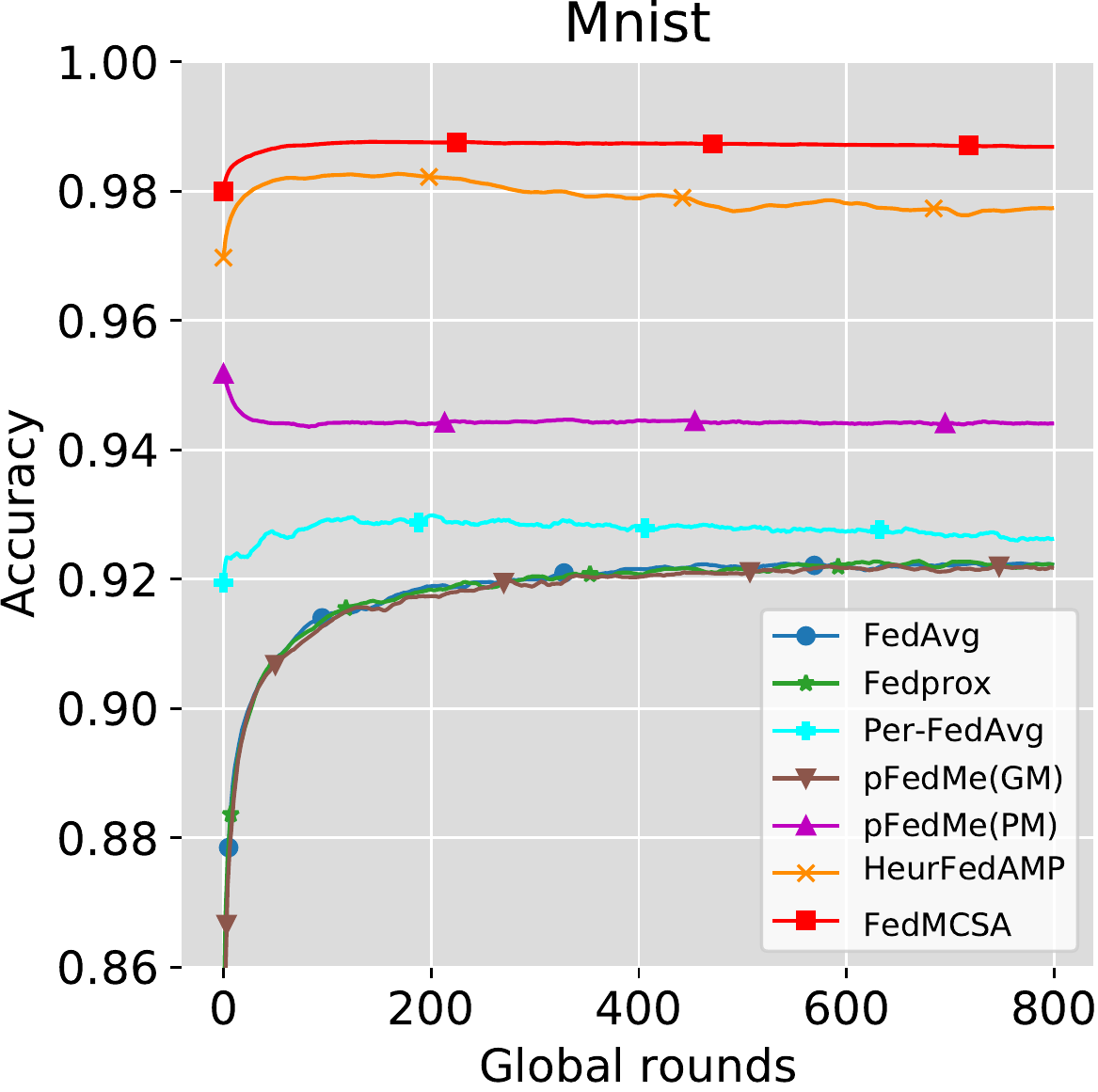}
	}
	\subfigure[]{
		\includegraphics[width=0.35\columnwidth]{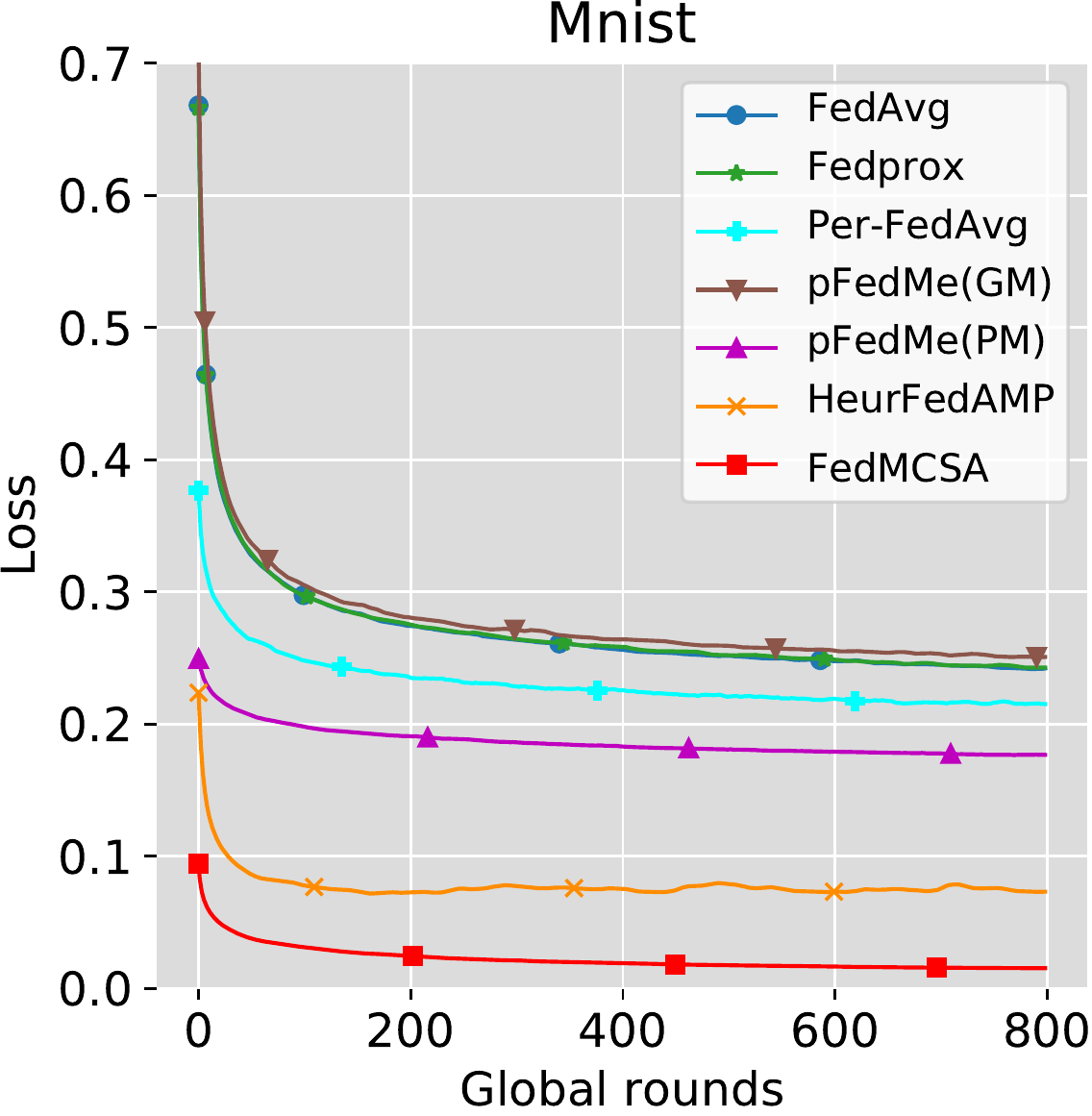}
	}
	\caption{Performance comparisons of FedAvg, Fedprox, Per-FedAvg, pFedMe(GM), pFedMe(PM), HeurFedAMP, and FedMCSA using MLR model on Synthetic and Mnist datasets.}
	\label{fig.exp.new.3}
\end{figure}
\begin{figure}[t]
	\centering
	\subfigure[]{
		\includegraphics[width=0.35\columnwidth]{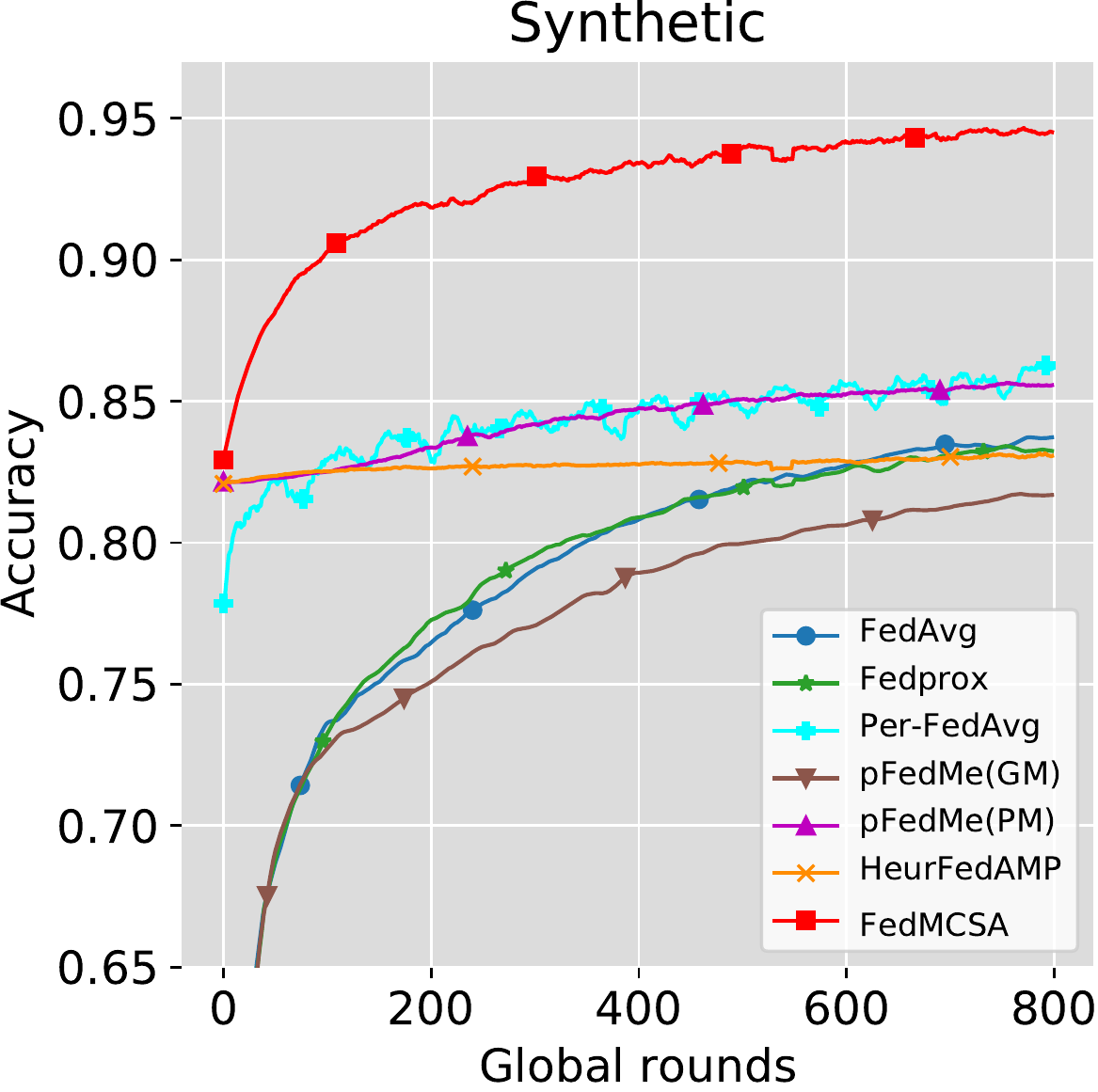}
	}
	\subfigure[]{
		\includegraphics[width=0.35\columnwidth]{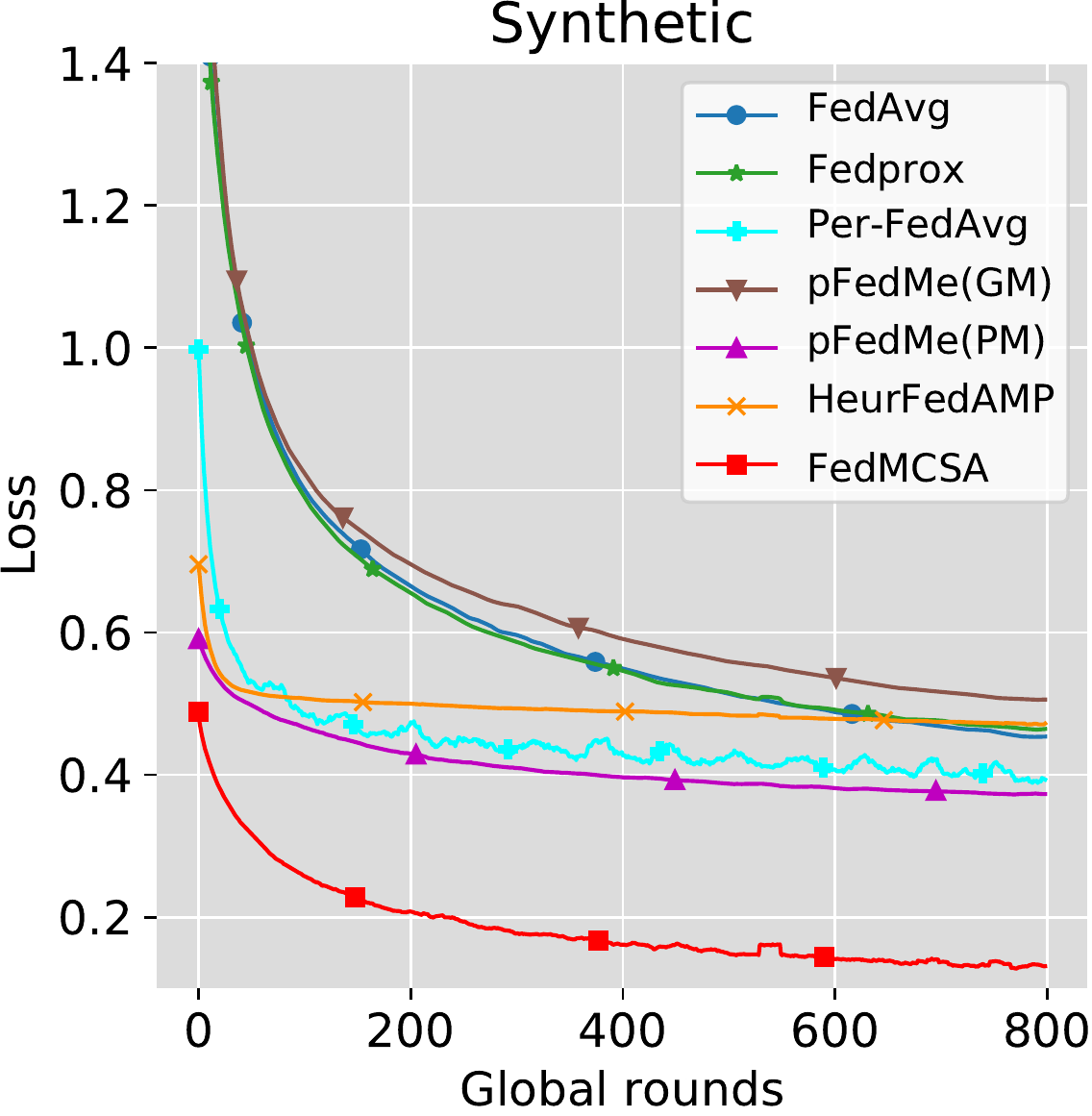}
	}
	\subfigure[]{
		\includegraphics[width=0.35\columnwidth]{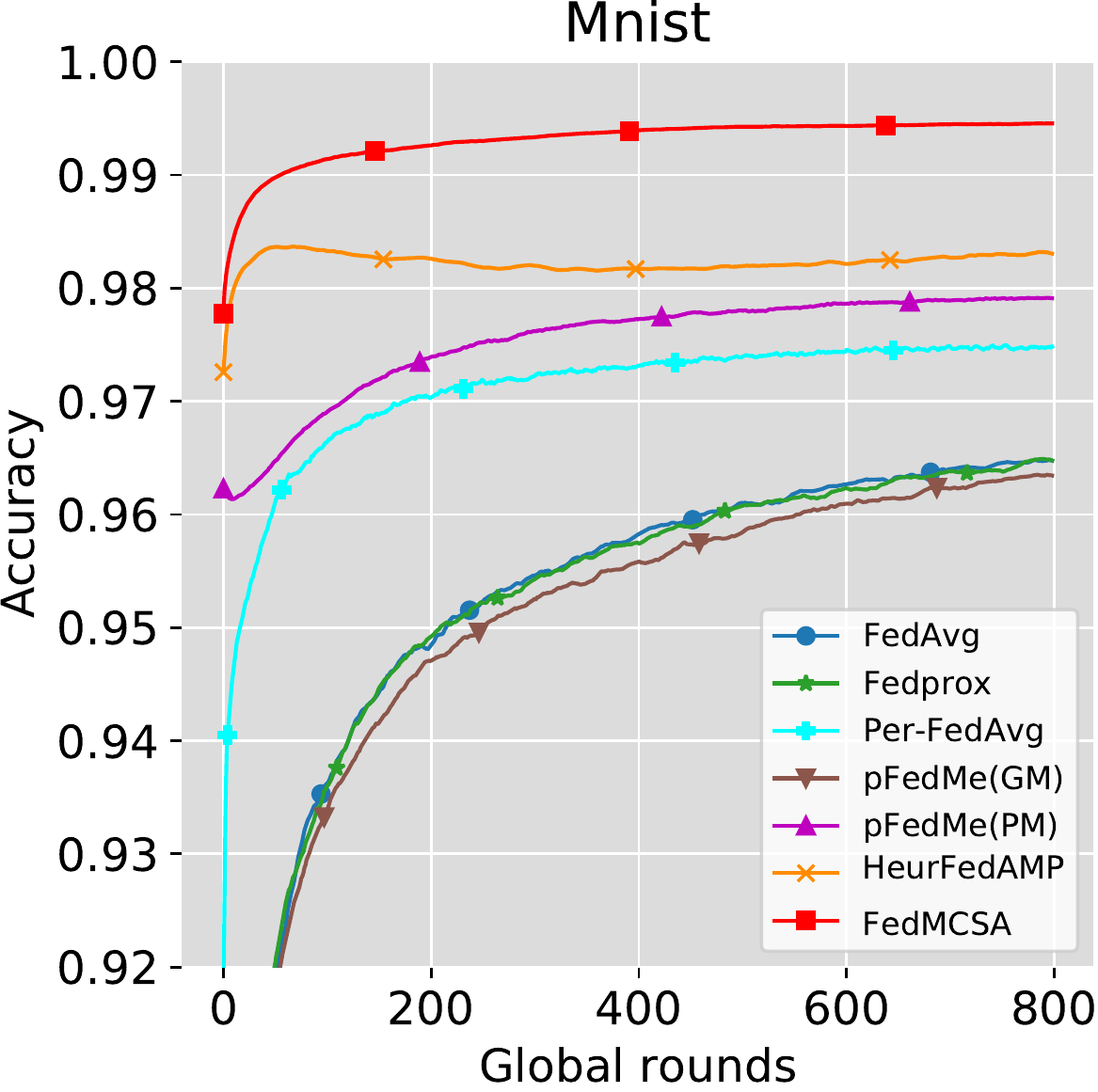}
	}
	\subfigure[]{
		\includegraphics[width=0.35\columnwidth]{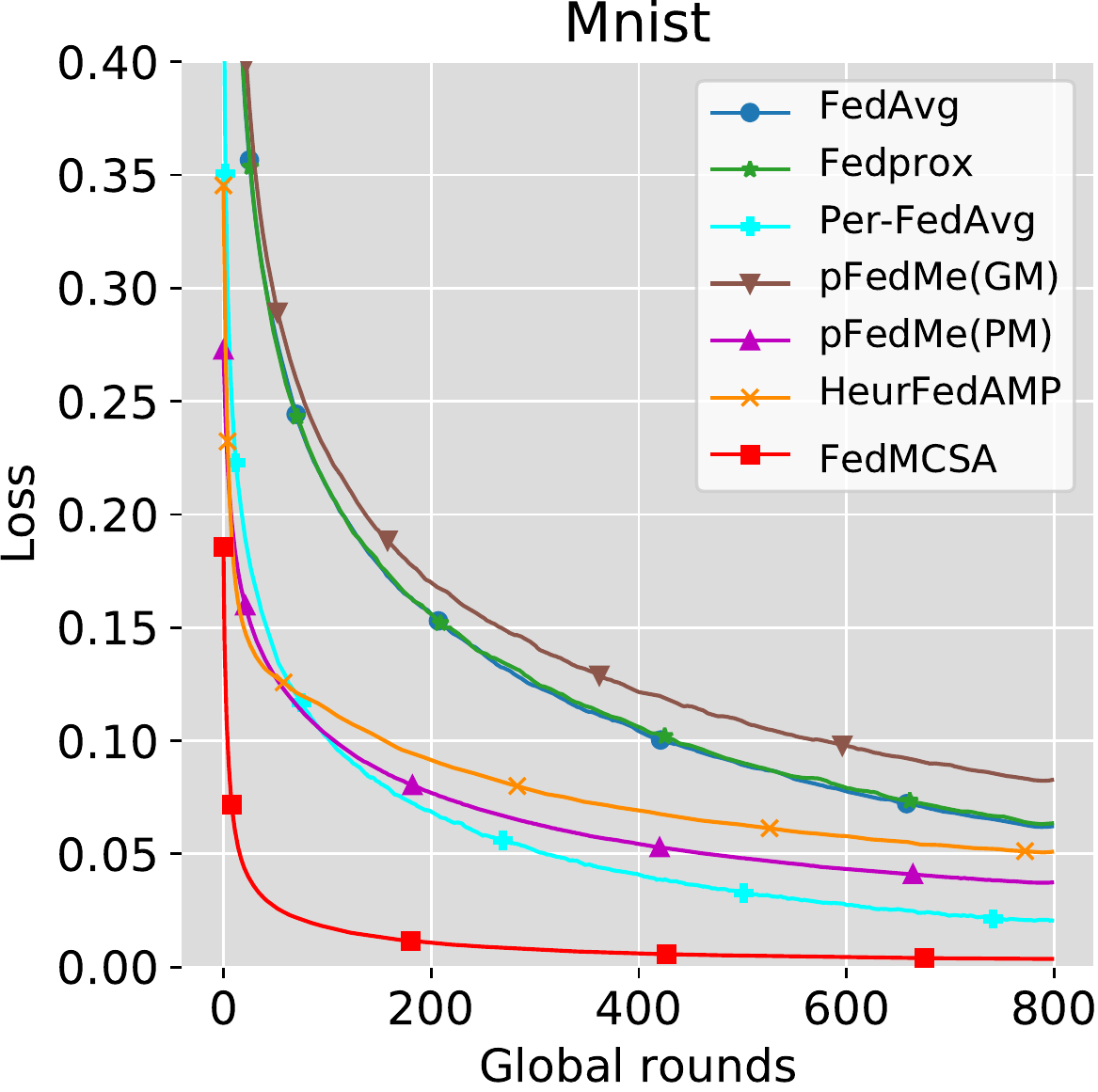}
	}
	\caption{Performance comparisons of FedAvg, Fedprox, Per-FedAvg, pFedMe(GM), pFedMe(PM), HeurFedAMP, and FedMCSA using DNN model on Synthetic and Mnist datasets.}
	\label{fig.exp.new.4}
\end{figure}

The comparisons for the same hyperparameters are shown in \figurename~\ref{fig.exp.new.3} and \figurename~\ref{fig.exp.new.4}. According to \figurename~\ref{fig.exp.new.3} and \figurename~\ref{fig.exp.new.4}, we can observe that FedMCSA achieves the best performance compared to other algorithms in different settings. Specifically, for the MLR model, FedMCSA is about 16.86$\%$, 16.66$\%$, 9.55$\%$, 17.57$\%$, 10.53$\%$and 9.64$\%$ more accurate than FedAvg, Fedprox, Per-FedAvg, pFedMe(GM), pFedMe(PM), and HeurFedAMP on Synthetic dataset, and 6.47$\%$, 6.42$\%$, 5.60$\%$, 6.53$\%$, 2.75$\%$, and 0.50$\%$ more accurate than those on Mnist dataset. For the DNN model, FedMCSA is about 11.39$\%$, 11.69$\%$, 7.52$\%$, 13.42$\%$, 9.07$\%$, and 11.61$\%$ more accurate than FedAvg, Fedprox, Per-FedAvg, pFedMe(GM), pFedMe(PM), and HeurFedAMP on Synthetic dataset, and 2.94$\%$, 2.96$\%$, 1.81$\%$, 3.06$\%$, 1.50$\%$, and 1.07$\%$ more accurate than those on Mnist dataset. 

It can be observed that the performance of FedAvg, Fedprox, and pFedMe(GM) is similar and significantly lower. The performance of Per-FedAvg, pFedMe(PM) and HeurFedAMP is similar and have advantages over FedAvg, Fedprox, and pFedMe(GM). Additionally, our proposed FedMCSA clearly surpasses PerFedAvg, pFedMe(PM), and HeurFedAMP, demonstrating its superiority. Although the optimization method of FedMCSA in the client is generally universal, with being similar to HeurFedAMP, FedMCSA is superior to FedAvg, Fedprox, pFedMe(GM), Per-FedAvg, pFedMe(PM), and HeurFedAMP in different settings. This is due to the fact that FedMCSA adopts the model components self-attention mechanism on the server side, which considers the significance of the internal components of the model from a novel perspective, and further implements an adaptive update of model components through parallel model components self-attention mechanisms. FedMCSA realizes the adaptive update of the entire model through the continuous adaptive update of model components. Consequently, the performance of FedMCSA is superior to FedAvg, Fedprox, Per-FedAvg, pFedMe, and HeurFedAMP under different models and datasets when the same hyperparameters are used.  

\subsubsection{The comparisons for fine-tuned hyperparameters.}
\begin{table}[t]
	\caption{Performance comparisons using fine-tuned hyperparameters. $|\mathcal{D}|$ = 20, $R$ = 20, and $T$ = 800 are fixed for all experiments and all results are averaged over 3 runs. (The best result is marked in bold).}
	\resizebox{\textwidth}{!}{
		\begin{tabular}{lllllllll}
			\toprule
			\multicolumn{1}{c}{\multirow{2}{*}{Algorithm}} &
			\multicolumn{2}{c}{Synthetic} &
			\multicolumn{2}{c}{Mnist} &
			\multicolumn{2}{c}{FMnist} &
			\multicolumn{2}{c}{Cifar10} \\ \cline{2-9} 
			\multicolumn{1}{c}{} &
			\multicolumn{1}{c}{MLR} &
			\multicolumn{1}{c}{DNN} &
			\multicolumn{1}{c}{MLR} &
			\multicolumn{1}{c}{DNN} &
			\multicolumn{1}{c}{MLR} &
			\multicolumn{1}{c}{DNN} &
			\multicolumn{1}{c}{MLR} &
			\multicolumn{1}{c}{DNN} \\ \midrule
			FedAvg~\cite{McMahan_2017_AISTATS_FedAvg}     & 78.04±0.21 & 84.30±0.03 & 92.39±0.02 & 96.86±0.01 & 84.56±0.01 & 85.80±0.04 & 38.43±0.25 & 46.10±0.14 \\
			Fedprox~\cite{TianLi_2020_Mlsys_FL}      & 77.51±0.22 & 84.01±0.06 & 92.42±0.03 & 96.89±0.02 & 84.45±0.01 & 85.68±0.10 & 38.35±0.41 & 45.92±0.16 \\
			Per-FedAvg~\cite{AlirezaFallah_2020_NIPS_PFL} & 85.15±0.06 & 88.50±0.07 & 94.18±0.05 & 98.15±0.01 & 98.82±0.04 & 99.31±0.01 & 57.79±0.08 & 81.64±0.09 \\
			pFedMe(GM)~\cite{CanhTDinh_2020_NIPS_PFL} & 76.37±0.18 & 82.70±0.28 & 92.08±0.01 & 96.81±0.03 & 83.55±0.11 & 84.30±0.12 & 36.21±0.11 & 46.89±0.39 \\
			pFedMe(PM)~\cite{CanhTDinh_2020_NIPS_PFL} & 84.97±0.08 & 87.01±0.08 & 97.24±0.01 & 98.93±0.01 & 99.20±0.01 & 99.30±0.01 & 66.70±0.09 & 84.05±0.13 \\
			HeurFedAMP~\cite{YutaoHuang_2021_AAAI_PFL}    & 84.42±0.05 & 84.00±0.06 & 98.33±0.03 & 98.42±0.01 & 99.19±0.01 & 99.13±0.02 & 72.18±0.03 & 82.41±0.03 \\
			\textbf{FedMCSA(Our)}                     & \textbf{95.27±0.01} & \textbf{96.26±0.02} & \textbf{98.87±0.01} & \textbf{99.58±0.01} & \textbf{99.33±0.01} & \textbf{99.39±0.01} & \textbf{75.91±0.02} & \textbf{84.40±0.05} \\ \bottomrule
	\end{tabular}}
	\label{table.exp.new.0}
\end{table}
The highest accuracy comparison results achieved by fine-tuning the hyperparameters of all algorithms are shown in \tablename~\ref{table.exp.new.0}.  In all settings, FedMCSA surpasses other algorithms to achieve the highest accuracy. Specifically, FedMCSA achieves 95.27\%, 99.87\%, 99.33\%, and 75.91\% on Synthetic, FMnist, Mnist, and Cifar10 datasets when using the MLR model. And FedMCSA achieves 96.26\%, 99.58\%, 99.39\%, and 84.40\% on Synthetic, FMnist, Mnist, and Cifar10 datasets when using the DNN model. 

As for the MLR model, among all the baseline algorithms, the best performances achieved on Synthetic, Mnist, FMnist, and Cifar10 datasets are 85.15\% of Per-FedAvg, 98.33\% of HeurFedAMP, and 99.20\% of pFedMe(PM), and 72.18\% of HeurFedAMP, respectively. As for the DNN model, among all the baseline algorithms, the best performances achieved on Synthetic, Mnist, FMnist, and Cifar10 datasets are 88.50\% of Per-FedAvg, 98.93\% of pFedMe(PM), and 99.31\% of Per-FedAvg, and 84.05\% of pFedMe(PM), respectively. It is evident that none of all baselines achieve the best accuracies using different models under different datasets simultaneously, whereas FedMCSA achieves the highest accuracies using both MLR and DNN models across the four datasets simultaneously.

From \figurename~\ref{fig.exp.new.3}, \figurename~\ref{fig.exp.new.4}, and \tablename~\ref{table.exp.new.0}, we can observe that the performance of the DNN model tends to be better than that of the MLR model under the same experimental conditions. This is because DNN model has a hidden layer with stronger information extraction ability than MLR model, and can more fully mine the information in clients' private data. 

The key to the superiority of FedMCSA is that it abandons the setting of a complete model as a basic unit to carry out the cooperation between different models, and treats each layer of the model as a basic unit from the perspective of finer granularity. By taking each layer's significance into consideration, FedMCSA explores the potential of the model components to make collaboration between different clients more targeted and effective. As a result, each model component can learn as much useful knowledge as possible from other models. With the advantage of model components self-attention mechanism, FedMCSA has achieved surprising performance using different models on Synthetic, Mnist, FMnist, and Cifar10 datasets.

\subsection{Ablation Studies}
To further analyze the model components self-attention mechanism of FedMCSA, we conduct ablation studies to verify the effectiveness of the mechanism from two different perspectives. 
On the one hand, we construct a variant of FedMCSA by replacing the model components self-attention mechanism with the average aggregation method. On the other hand, we integrate this mechanism into the most basic FL algorithm, FedAvg, as well as the representative personalized FL with average aggregation, pFedMe. Considering that the performance of the personalized model pFedMe(PM) is always better than the global model pFedMe(GM) in the same setting, here we use pFedMe(PM) to represent the pFedMe. Note that in each group of experiments, the settings are identical except for the variables of interest. 

Specifically, we conduct three variants of FedMCSA, FedAvg, and pFedMe as follows.

\begin{itemize}
	\item FedMCSA$^{-M\!C\!S\!A}$: The variant model that replaces the model components self-attention mechanism with the average aggregation method.
	
	\item FedAvg$^{+M\!C\!S\!A}$: The variant model that replaces FedAvg's aggregation method with the model components self-attention mechanism.
	
	\item pFedMe$^{+M\!C\!S\!A}$: The variant model that replaces pFedMe(PM)'s  aggregation method with the model components self-attention mechanism.
\end{itemize}

\begin{table}[t]
	\caption{Accuracy on the variants of FedMCSA, FedAvg, and pFedMe on four benchmark datasets. $"\Delta"$ represents the change values of the variants compared with FedMCSA, FedAvg, and pFedMe and all results are averaged over 3 runs.}
	\resizebox{\textwidth}{!}{
	\begin{tabular}{lllllllll}
		\toprule
		&
		\multicolumn{2}{c}{Synthetic} &
		\multicolumn{2}{c}{Mnist} &
		\multicolumn{2}{c}{FMnist} &
		\multicolumn{2}{c}{Cifar10} \\ \cline{2-9} 
		&
		\multicolumn{1}{c}{MLR} &
		\multicolumn{1}{c}{DNN} &
		\multicolumn{1}{c}{MLR} &
		\multicolumn{1}{c}{DNN} &
		\multicolumn{1}{c}{MLR} &
		\multicolumn{1}{c}{DNN} &
		\multicolumn{1}{c}{MLR} &
		\multicolumn{1}{c}{DNN} \\ \midrule
		FedMCSA             & 94.06\% & 95.26\% & 98.82\% & 99.48\% & 99.32\% & 99.37\% & 72.18\% & 84.24\% \\ \hline
		FedMCSA$^{-M\!C\!S\!A}$   & 76.35\% & 83.08\% & 92.20\% & 96.84\% & 83.76\% & 84.42\% & 34.05\% & 46.26\% \\
		$\Delta$            & 17.71\%$\downarrow$ & 12.18\%$\downarrow$ & 6.62\%$\downarrow$  & 2.64\%$\downarrow$  & 15.56\%$\downarrow$ & 14.95\%$\downarrow$ & 38.13\%$\downarrow$ & 37.98\%$\downarrow$ \\ \hline
		FedAvg              & 77.26\% & 83.55\% & 92.39\% & 96.54\% & 84.48\% & 83.87\% & 34.81\% & 45.60\% \\ \hline
		FedAvg$^{+M\!C\!S\!A}$ & 94.32\% & 95.52\% & 98.83\% & 99.49\% & 99.31\% & 99.36\% & 72.13\% & 84.37\% \\
		$\Delta$            & 17.06\%$\uparrow$ & 11.97\%$\uparrow$ & 6.44\%$\uparrow$  & 2.95\%$\uparrow$  & 14.83\%$\uparrow$ & 15.49\%$\uparrow$ & 37.32\%$\uparrow$ & 38.77\%$\uparrow$ \\ \hline
		pFedMe              & 83.56\% & 86.04\% & 96.06\% & 97.98\% & 98.95\% & 99.23\% & 65.46\% & 83.77\% \\ \hline
		pFedMe$^{+M\!C\!S\!A}$ & 93.86\% & 93.49\% & 98.85\% & 99.48\% & 99.33\% & 99.34\% & 74.62\% & 84.70\% \\
		$\Delta$            & 10.30\%$\uparrow$ & 7.45\%$\uparrow$  & 2.79\%$\uparrow$  & 1.50\%$\uparrow$  & 0.38\%$\uparrow$  & 0.11\%$\uparrow$  & 9.16\%$\uparrow$  & 0.93\%$\uparrow$  \\ \bottomrule
	\end{tabular}}
\label{table.exp.ablation}
\end{table} 

The experimental results of ablation studies are shown in \tablename~\ref{table.exp.ablation}. Compared with FedMCSA, FedMCSA$^{-M\!C\!S\!A}$ has a significant drop in accuracy on the four benchmark datasets whether using the MLR model or the DNN model. FedMCSA$^{-M\!C\!S\!A}$ achieves relatively good performance on Mnist dataset, with 6.62\% and 2.64\% drop in accuracy compared to FedMCSA using the MLR and DNN models, respectively. This is because the Mnist dataset is relatively simple, and the basic average aggregation method can mine sufficient effective knowledge. When the dataset is relatively complex, e.g., the Synthetic dataset, the impact of the absence of the model components self-attention mechanism is significantly enlarged. It is noted that compared with FedMCSA, FedMCSA$^{-M\!C\!S\!A}$ drops by 38.13\% with the MLR model and 37.98\% with the DNN model on Cifar10 dataset. The consistent accuracy drop across different datasets illustrates the effectiveness of the model components self-attention mechanism of FedMCSA, and the greater accuracy impact on more complex datasets further illustrates that the role of the model components self-attention mechanism is more pronounced on relatively complex datasets.

We also compare the performance of FedAvg, FedAvg$^{+M\!C\!S\!A}$, pFedMe, pFedMe$^{+M\!C\!S\!A}$ on the four benchmark datasets.
From \tablename~\ref{table.exp.ablation}, we can observe that the model components self-attention mechanism of FedMCSA not only achieves the overall improvement on FedAvg, but also achieves the improvement on pFedMe. FedAvg$^{+M\!C\!S\!A}$ has a significant accuracy improvement over FedAvg on the four benchmark datasets. Especially on Cifar10 dataset, compared with FedAvg, FedAvg$^{+M\!C\!S\!A}$ has improved the accuracy by 37.32\% and 38.77\% under the MLR model and DNN model, respectively. 
Although pFedMe has achieved good performance on different datasets, pFedMe$^{+M\!C\!S\!A}$ still achieves performance improvements under a variety of different settings. On Synthetic dataset, pFedMe$^{+M\!C\!S\!A}$ improves the accuracy by 10.3\% and 7.45\% under the MLR model and DNN model, respectively.

The results of \tablename~\ref{table.exp.ablation} demonstrate the effectiveness of the model components self-attention mechanism of FedMCSA from the consistent accuracy decrease of FedMCSA$^{-M\!C\!S\!A}$ and the consistent accuracy improvement of FedAvg$^{+M\!C\!S\!A}$ and pFedMe$^{+M\!C\!S\!A}$. Furthermore, the performance improvement across different datasets, different network models, and different algorithms further illustrates the generality and usability of the mechanism.

\begin{figure}[t]
	\centering
	\subfigure[]{
		\includegraphics[width=0.22\columnwidth]{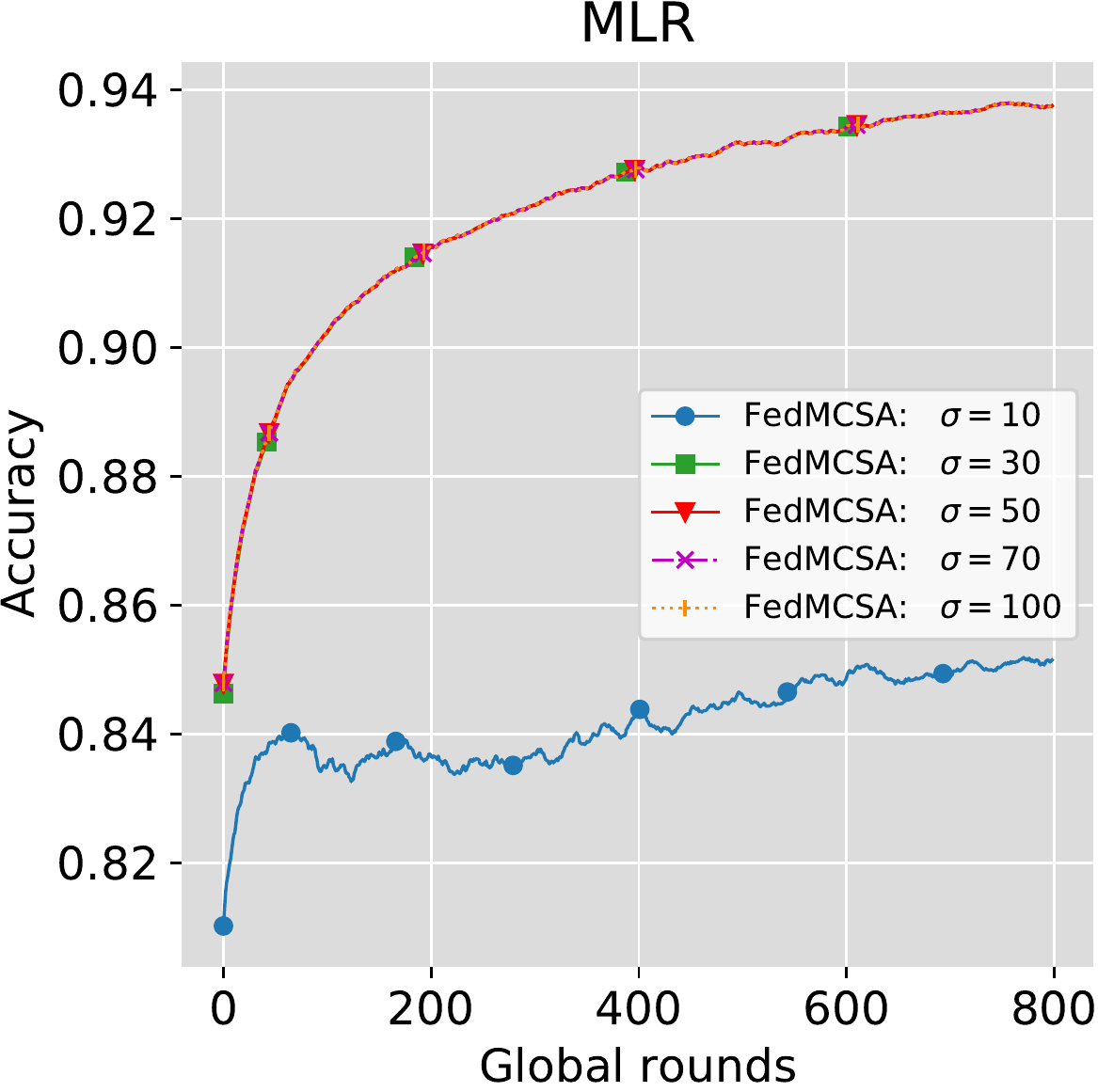}
	}
	\subfigure[]{
		\includegraphics[width=0.22\columnwidth]{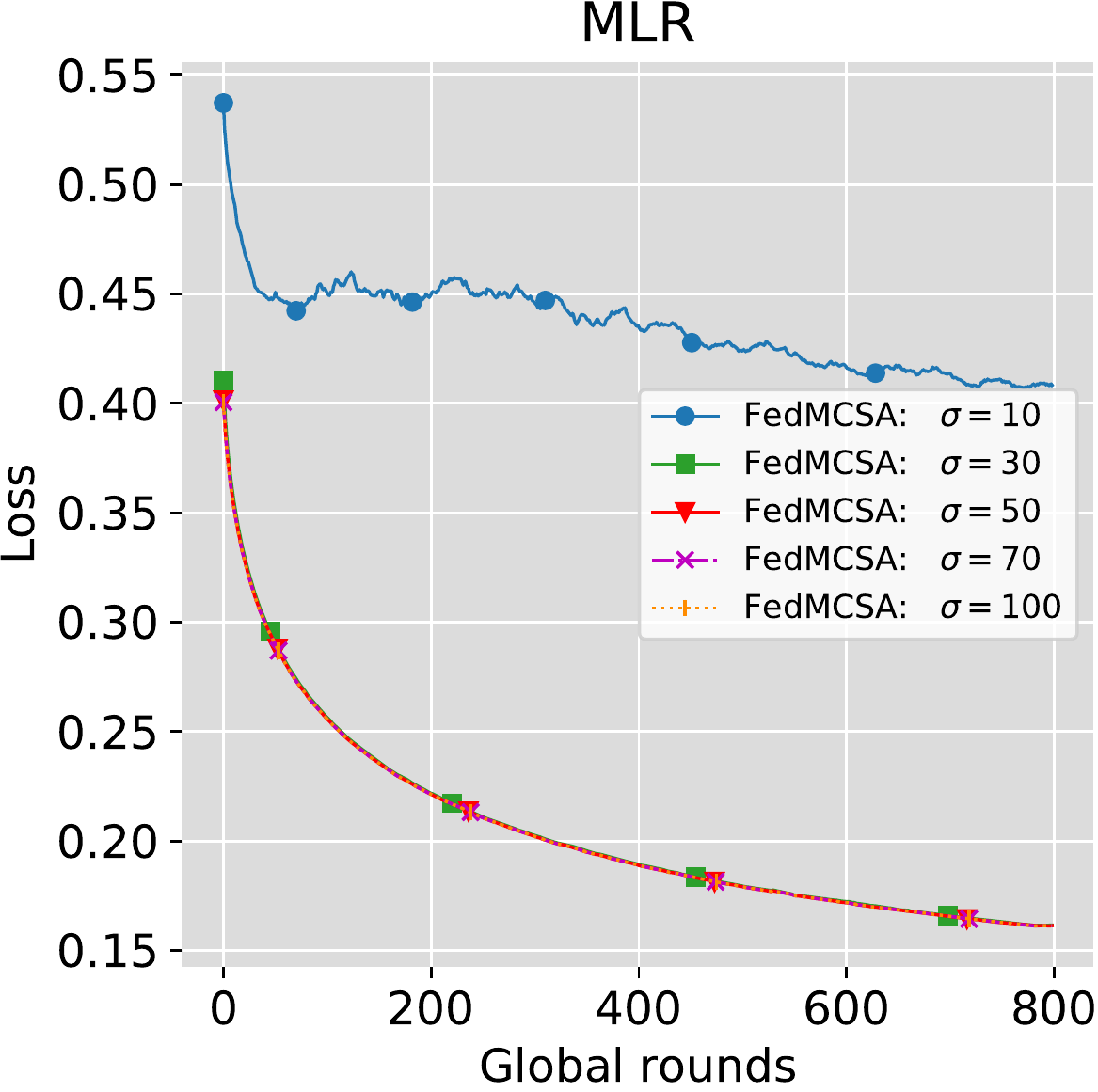}
	}
	\subfigure[]{
		\includegraphics[width=0.22\columnwidth]{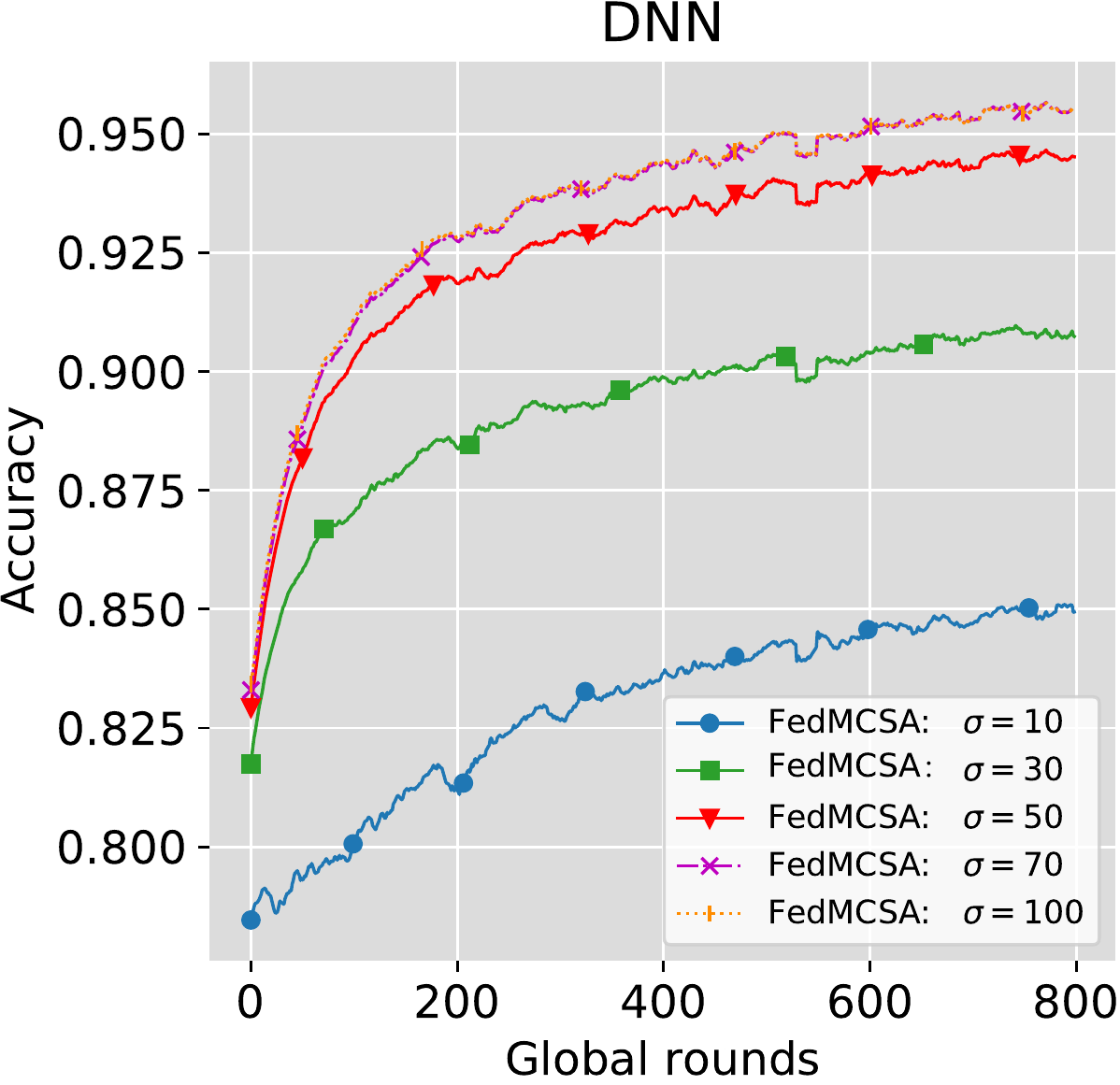}
	}
	\subfigure[]{
		\includegraphics[width=0.21\columnwidth]{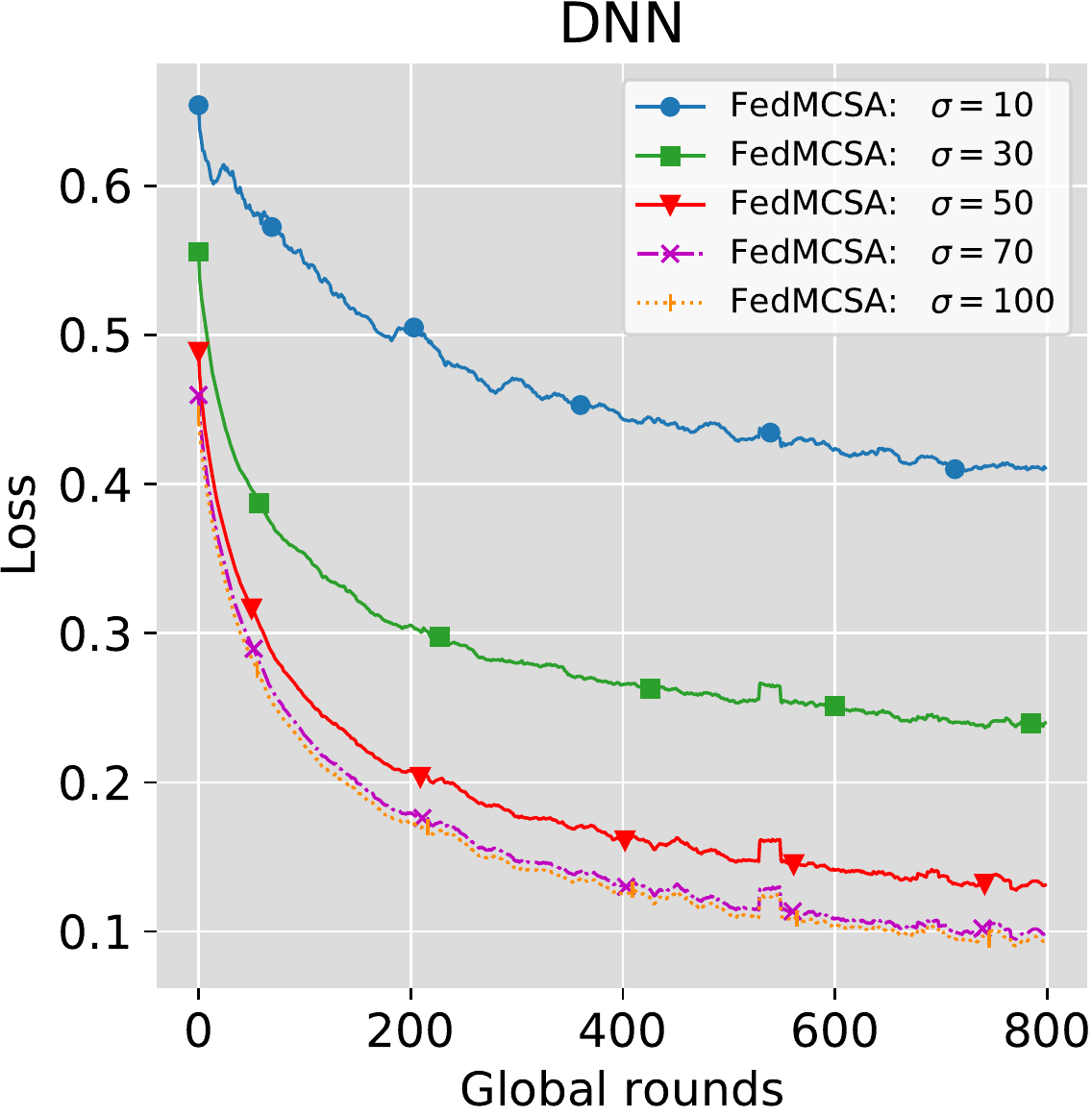}
	}
	\caption{Impact of $\sigma$ on the convergence of FedMCSA using MLR and DNN models on Synthetic dataset.}
	\label{fig.exp.new.0}
\end{figure}
\subsection{Impact of The Important Hyperparameter}
To investigate the impact of the important hyperparameter $\sigma$ on the convergence of FedMCSA, we conduct diverse experiments on the Synthetic dataset using both MLR and DNN models. 

As $\sigma$ is the scale of model components similarity in the FedMCSA, $\sigma$ is considered as a hyperparameter of FedMCSA. As illustrated in \figurename~\ref{fig.exp.new.0}, all the parameters except $\sigma$ are fixed during the experiments. For MLR, once $\sigma$ increases to 30, the upper limit of FedMCSA's accuracy is reached under the current conditions, while the upper limit of accuracy for DNN is only realized when $\sigma=70$. We observe that FedMCSA requires a moderate value to compute the personalized model. When the value of $\sigma$ is too small, FedMCSA will converge slowly and reduce the accuracy. However, when the value of $\sigma$ exceeds a certain threshold, the accuracy of FedMCSA is not further improved, while an excessive value of $\sigma$ will cause the computational cost of the process to increase sharply. As a result, the value of $\sigma$ = 50 is selected as the default in the case of both MLR and DNN for the Synthetic dataset. Similarly, the value of $\sigma$ = 50 is selected as the default in the case of both MLR and DNN for the Mnist, FMnist, and Cifar10 datasets.

\section{Conclusion}
In this paper, we propose a new framework, federated model components self-attention, to handle the Non-IID data in FL, which devise the model components self-attention mechanism from a novel perspective of the component-based aggregation method.
Different from the previous personalized FL approaches that treat the entire model as a basic unit for aggregation, FedMCSA promotes purposeful and refined collaboration among clients through parallel model components self-attention mechanism. 
Our extensive experiments on benchmark datasets demonstrate the superior performance of FedMCSA compared with FedAvg, Fedprox, Per-FedAvg, pFedMe, and HeurFedAMP in different settings. 
Moreover, the empirical experiment results demonstrate that the model components self-attention mechanism of FedMCSA has effectiveness and universality, and can also be easily integrated into existing personalized FL to further improve their performance.
We hope that future research of personalized FL will not only focus on external optimization of the model, but also pay more attention to exploring the potential of collaboration between model components, which could lead to a more superior FL.


\bibliography{tempreferences}

\begin{thebibliography}{10}
\expandafter\ifx\csname url\endcsname\relax
  \def\url#1{\texttt{#1}}\fi
\expandafter\ifx\csname urlprefix\endcsname\relax\def\urlprefix{URL }\fi
\expandafter\ifx\csname href\endcsname\relax
  \def\href#1#2{#2} \def\path#1{#1}\fi

\bibitem{McMahan_2017_AISTATS_FedAvg}
B.~McMahan, E.~Moore, D.~Ramage, S.~Hampson, B.~A. y~Arcas,
  \href{http://proceedings.mlr.press/v54/mcmahan17a.html}{Communication-efficient
  learning of deep networks from decentralized data}, in: Proceedings of the
  20th International Conference on Artificial Intelligence and Statistics,
  {AISTATS} 2017, 20-22 April 2017, Fort Lauderdale, FL, {USA}, Vol.~54 of
  Proceedings of Machine Learning Research, {PMLR}, 2017, pp. 1273--1282.
\newline\urlprefix\url{http://proceedings.mlr.press/v54/mcmahan17a.html}

\bibitem{QiangYang_2019_ACMTrans_FLsurvey}
Q.~Yang, Y.~Liu, T.~Chen, Y.~Tong,
  \href{https://doi.org/10.1145/3298981}{Federated machine learning: Concept
  and applications}, {ACM} Trans. Intell. Syst. Technol. 10~(2) (2019)
  12:1--12:19.
\newblock \href {http://dx.doi.org/10.1145/3298981}
  {\path{doi:10.1145/3298981}}.
\newline\urlprefix\url{https://doi.org/10.1145/3298981}

\bibitem{TianLi_2020_SignalProcess_FLsurvey}
T.~Li, A.~K. Sahu, A.~Talwalkar, V.~Smith,
  \href{https://doi.org/10.1109/MSP.2020.2975749}{Federated learning:
  Challenges, methods, and future directions}, {IEEE} Signal Process. Mag.
  37~(3) (2020) 50--60.
\newblock \href {http://dx.doi.org/10.1109/MSP.2020.2975749}
  {\path{doi:10.1109/MSP.2020.2975749}}.
\newline\urlprefix\url{https://doi.org/10.1109/MSP.2020.2975749}

\bibitem{AndrewHard_2018_arXiv_FL}
A.~Hard, K.~Rao, R.~Mathews, S.~Ramaswamy, F.~Beaufays, S.~Augenstein,
  H.~Eichner, C.~Kiddon, D.~Ramage, Federated learning for mobile keyboard
  prediction, arXiv preprint arXiv:1811.03604.

\bibitem{VirajKulkarni_2020_arXiv_PFL}
V.~Kulkarni, M.~Kulkarni, A.~Pant, Survey of personalization techniques for
  federated learning, in: 2020 Fourth World Conference on Smart Trends in
  Systems, Security and Sustainability (WorldS4), IEEE, 2020, pp. 794--797.

\bibitem{wang2019federated}
K.~Wang, R.~Mathews, C.~Kiddon, H.~Eichner, F.~Beaufays, D.~Ramage,
  \href{http://arxiv.org/abs/1910.10252}{Federated evaluation of on-device
  personalization}, CoRR abs/1910.10252.
\newblock \href {http://arxiv.org/abs/1910.10252} {\path{arXiv:1910.10252}}.
\newline\urlprefix\url{http://arxiv.org/abs/1910.10252}

\bibitem{mansour2020three}
Y.~Mansour, M.~Mohri, J.~Ro, A.~T. Suresh,
  \href{https://arxiv.org/abs/2002.10619}{Three approaches for personalization
  with applications to federated learning}, CoRR abs/2002.10619.
\newblock \href {http://arxiv.org/abs/2002.10619} {\path{arXiv:2002.10619}}.
\newline\urlprefix\url{https://arxiv.org/abs/2002.10619}

\bibitem{AlirezaFallah_2020_NIPS_PFL}
A.~Fallah, A.~Mokhtari, A.~E. Ozdaglar,
  \href{https://proceedings.neurips.cc/paper/2020/hash/24389bfe4fe2eba8bf9aa9203a44cdad-Abstract.html}{Personalized
  federated learning with theoretical guarantees: {A} model-agnostic
  meta-learning approach}, in: Advances in Neural Information Processing
  Systems 33: Annual Conference on Neural Information Processing Systems 2020,
  NeurIPS 2020, December 6-12, 2020, virtual, 2020.
\newline\urlprefix\url{https://proceedings.neurips.cc/paper/2020/hash/24389bfe4fe2eba8bf9aa9203a44cdad-Abstract.html}

\bibitem{CanhTDinh_2020_NIPS_PFL}
C.~T. Dinh, N.~H. Tran, T.~D. Nguyen,
  \href{https://proceedings.neurips.cc/paper/2020/hash/f4f1f13c8289ac1b1ee0ff176b56fc60-Abstract.html}{Personalized
  federated learning with moreau envelopes}, in: H.~Larochelle, M.~Ranzato,
  R.~Hadsell, M.~Balcan, H.~Lin (Eds.), Advances in Neural Information
  Processing Systems 33: Annual Conference on Neural Information Processing
  Systems 2020, NeurIPS 2020, December 6-12, 2020, virtual, 2020.
\newline\urlprefix\url{https://proceedings.neurips.cc/paper/2020/hash/f4f1f13c8289ac1b1ee0ff176b56fc60-Abstract.html}

\bibitem{MichaelZhang_2021_ICLR_PFL}
M.~Zhang, K.~Sapra, S.~Fidler, S.~Yeung, J.~M. Alvarez,
  \href{https://openreview.net/forum?id=ehJqJQk9cw}{Personalized federated
  learning with first order model optimization}, in: 9th International
  Conference on Learning Representations, {ICLR} 2021, Virtual Event, Austria,
  May 3-7, 2021, OpenReview.net, 2021.
\newline\urlprefix\url{https://openreview.net/forum?id=ehJqJQk9cw}

\bibitem{YutaoHuang_2021_AAAI_PFL}
Y.~Huang, L.~Chu, Z.~Zhou, L.~Wang, J.~Liu, J.~Pei, Y.~Zhang,
  \href{https://ojs.aaai.org/index.php/AAAI/article/view/16960}{Personalized
  cross-silo federated learning on non-iid data}, in: Thirty-Fifth {AAAI}
  Conference on Artificial Intelligence, {AAAI} 2021, Thirty-Third Conference
  on Innovative Applications of Artificial Intelligence, {IAAI} 2021, The
  Eleventh Symposium on Educational Advances in Artificial Intelligence, {EAAI}
  2021, Virtual Event, February 2-9, 2021, {AAAI} Press, 2021, pp. 7865--7873.
\newline\urlprefix\url{https://ojs.aaai.org/index.php/AAAI/article/view/16960}

\bibitem{FilipHanzely_2020_NIPS_PFL}
F.~Hanzely, S.~Hanzely, S.~Horv{\'{a}}th, P.~Richt{\'{a}}rik,
  \href{https://proceedings.neurips.cc/paper/2020/hash/187acf7982f3c169b3075132380986e4-Abstract.html}{Lower
  bounds and optimal algorithms for personalized federated learning}, in:
  Advances in Neural Information Processing Systems 33: Annual Conference on
  Neural Information Processing Systems 2020, NeurIPS 2020, December 6-12,
  2020, virtual, 2020.
\newline\urlprefix\url{https://proceedings.neurips.cc/paper/2020/hash/187acf7982f3c169b3075132380986e4-Abstract.html}

\bibitem{verleysen2005curse}
M.~Verleysen, D.~Fran{\c{c}}ois,
  \href{https://doi.org/10.1007/11494669\_93}{The curse of dimensionality in
  data mining and time series prediction}, in: J.~Cabestany, A.~Prieto, F.~S.
  Hern{\'{a}}ndez (Eds.), Computational Intelligence and Bioinspired Systems,
  8th International Work-Conference on Artificial Neural Networks, {IWANN}
  2005, Vilanova i la Geltr{\'{u}}, Barcelona, Spain, June 8-10, 2005,
  Proceedings, Vol. 3512 of Lecture Notes in Computer Science, Springer, 2005,
  pp. 758--770.
\newblock \href {http://dx.doi.org/10.1007/11494669\_93}
  {\path{doi:10.1007/11494669\_93}}.
\newline\urlprefix\url{https://doi.org/10.1007/11494669\_93}

\bibitem{TianLi_2020_Mlsys_FL}
T.~Li, A.~K. Sahu, M.~Zaheer, M.~Sanjabi, A.~Talwalkar, V.~Smith,
  \href{https://proceedings.mlsys.org/book/316.pdf}{Federated optimization in
  heterogeneous networks}, in: Proceedings of Machine Learning and Systems
  2020, MLSys 2020, Austin, TX, USA, March 2-4, 2020, mlsys.org, 2020.
\newline\urlprefix\url{https://proceedings.mlsys.org/book/316.pdf}

\bibitem{LigengZhu_2019_NIPS_attack}
L.~Zhu, Z.~Liu, S.~Han,
  \href{https://proceedings.neurips.cc/paper/2019/hash/60a6c4002cc7b29142def8871531281a-Abstract.html}{Deep
  leakage from gradients}, in: Advances in Neural Information Processing
  Systems 32: Annual Conference on Neural Information Processing Systems 2019,
  NeurIPS 2019, December 8-14, 2019, Vancouver, BC, Canada, 2019, pp.
  14747--14756.
\newline\urlprefix\url{https://proceedings.neurips.cc/paper/2019/hash/60a6c4002cc7b29142def8871531281a-Abstract.html}

\bibitem{JinhyunSo_2020_NIPS_FLPrivacy}
J.~So, B.~G{\"{u}}ler, S.~Avestimehr,
  \href{https://proceedings.neurips.cc/paper/2020/hash/5bf8aaef51c6e0d363cbe554acaf3f20-Abstract.html}{A
  scalable approach for privacy-preserving collaborative machine learning}, in:
  Advances in Neural Information Processing Systems 33: Annual Conference on
  Neural Information Processing Systems 2020, NeurIPS 2020, December 6-12,
  2020, virtual, 2020.
\newline\urlprefix\url{https://proceedings.neurips.cc/paper/2020/hash/5bf8aaef51c6e0d363cbe554acaf3f20-Abstract.html}

\bibitem{StaceyTruex_2019_AISec_FLPrivacy}
S.~Truex, N.~Baracaldo, A.~Anwar, T.~Steinke, H.~Ludwig, R.~Zhang, Y.~Zhou,
  \href{https://doi.org/10.1145/3338501.3357370}{A hybrid approach to
  privacy-preserving federated learning}, in: Proceedings of the 12th {ACM}
  Workshop on Artificial Intelligence and Security, AISec@CCS 2019, London, UK,
  November 15, 2019, {ACM}, 2019, pp. 1--11.
\newblock \href {http://dx.doi.org/10.1145/3338501.3357370}
  {\path{doi:10.1145/3338501.3357370}}.
\newline\urlprefix\url{https://doi.org/10.1145/3338501.3357370}

\bibitem{AlekseiTriastcyn_2019_BigData_FLPrivacy}
A.~Triastcyn, B.~Faltings,
  \href{https://doi.org/10.1109/BigData47090.2019.9005465}{Federated learning
  with bayesian differential privacy}, in: C.~Baru, J.~Huan, L.~Khan, X.~Hu,
  R.~Ak, Y.~Tian, R.~S. Barga, C.~Zaniolo, K.~Lee, Y.~F. Ye (Eds.), 2019 {IEEE}
  International Conference on Big Data (Big Data), Los Angeles, CA, USA,
  December 9-12, 2019, {IEEE}, 2019, pp. 2587--2596.
\newblock \href {http://dx.doi.org/10.1109/BigData47090.2019.9005465}
  {\path{doi:10.1109/BigData47090.2019.9005465}}.
\newline\urlprefix\url{https://doi.org/10.1109/BigData47090.2019.9005465}

\bibitem{JennyHamer_2020_ICML_FLCommunication}
J.~Hamer, M.~Mohri, A.~T. Suresh,
  \href{http://proceedings.mlr.press/v119/hamer20a.html}{Fedboost: {A}
  communication-efficient algorithm for federated learning}, in: Proceedings of
  the 37th International Conference on Machine Learning, {ICML} 2020, 13-18
  July 2020, Virtual Event, Vol. 119 of Proceedings of Machine Learning
  Research, {PMLR}, 2020, pp. 3973--3983.
\newline\urlprefix\url{http://proceedings.mlr.press/v119/hamer20a.html}

\bibitem{DanielRothchild_2020_ICML_FLCommunication}
D.~Rothchild, A.~Panda, E.~Ullah, N.~Ivkin, I.~Stoica, V.~Braverman,
  J.~Gonzalez, R.~Arora,
  \href{http://proceedings.mlr.press/v119/rothchild20a.html}{Fetchsgd:
  Communication-efficient federated learning with sketching}, in: Proceedings
  of the 37th International Conference on Machine Learning, {ICML} 2020, 13-18
  July 2020, Virtual Event, Vol. 119 of Proceedings of Machine Learning
  Research, {PMLR}, 2020, pp. 8253--8265.
\newline\urlprefix\url{http://proceedings.mlr.press/v119/rothchild20a.html}

\bibitem{GrigoryMalinovskiy_2020_ICML_FLCommunication}
G.~Malinovskiy, D.~Kovalev, E.~Gasanov, L.~Condat, P.~Richt{\'{a}}rik,
  \href{http://proceedings.mlr.press/v119/malinovskiy20a.html}{From local {SGD}
  to local fixed-point methods for federated learning}, in: Proceedings of the
  37th International Conference on Machine Learning, {ICML} 2020, 13-18 July
  2020, Virtual Event, Vol. 119 of Proceedings of Machine Learning Research,
  {PMLR}, 2020, pp. 6692--6701.
\newline\urlprefix\url{http://proceedings.mlr.press/v119/malinovskiy20a.html}

\bibitem{RuiyuanWu_2021_AAAI_PFL}
R.~Wu, A.~Scaglione, H.~Wai, N.~Karako{\c{c}}, K.~Hreinsson, W.~Ma,
  \href{https://ojs.aaai.org/index.php/AAAI/article/view/17240}{Federated block
  coordinate descent scheme for learning global and personalized models}, in:
  Thirty-Fifth {AAAI} Conference on Artificial Intelligence, {AAAI} 2021,
  Thirty-Third Conference on Innovative Applications of Artificial
  Intelligence, {IAAI} 2021, The Eleventh Symposium on Educational Advances in
  Artificial Intelligence, {EAAI} 2021, Virtual Event, February 2-9, 2021,
  {AAAI} Press, 2021, pp. 10355--10362.
\newline\urlprefix\url{https://ojs.aaai.org/index.php/AAAI/article/view/17240}

\bibitem{JinzeWu_2021_WWW_PFL}
J.~Wu, Q.~Liu, Z.~Huang, Y.~Ning, H.~Wang, E.~Chen, J.~Yi, B.~Zhou,
  \href{https://doi.org/10.1145/3442381.3449926}{Hierarchical personalized
  federated learning for user modeling}, in: {WWW} '21: The Web Conference
  2021, Virtual Event / Ljubljana, Slovenia, April 19-23, 2021, {ACM} /
  {IW3C2}, 2021, pp. 957--968.
\newblock \href {http://dx.doi.org/10.1145/3442381.3449926}
  {\path{doi:10.1145/3442381.3449926}}.
\newline\urlprefix\url{https://doi.org/10.1145/3442381.3449926}

\bibitem{pmlr-v139-collins21a}
L.~Collins, H.~Hassani, A.~Mokhtari, S.~Shakkottai, Exploiting shared
  representations for personalized federated learning, in: Proceedings of the
  38th International Conference on Machine Learning, Vol. 139 of Proceedings of
  Machine Learning Research, PMLR, 2021, pp. 2089--2099.

\bibitem{pmlr-v139-shamsian21a}
A.~Shamsian, A.~Navon, E.~Fetaya, G.~Chechik, Personalized federated learning
  using hypernetworks, in: Proceedings of the 38th International Conference on
  Machine Learning, Vol. 139 of Proceedings of Machine Learning Research, PMLR,
  2021, pp. 9489--9502.

\bibitem{FilipHanzely_2020_arXiv_FLMixture}
F.~Hanzely, P.~Richt{\'{a}}rik,
  \href{https://arxiv.org/abs/2002.05516}{Federated learning of a mixture of
  global and local models}, arXiv preprint arXiv:2002.05516\href
  {http://arxiv.org/abs/2002.05516} {\path{arXiv:2002.05516}}.
\newline\urlprefix\url{https://arxiv.org/abs/2002.05516}

\bibitem{YuyangDeng_2020_arXiv_PFL}
Y.~Deng, M.~M. Kamani, M.~Mahdavi,
  \href{https://arxiv.org/abs/2003.13461}{Adaptive personalized federated
  learning}, arXiv preprint arXiv:2003.13461\href
  {http://arxiv.org/abs/2003.13461} {\path{arXiv:2003.13461}}.
\newline\urlprefix\url{https://arxiv.org/abs/2003.13461}

\bibitem{SenLin_2020_ICDCS_FLMetaLearning}
S.~Lin, G.~Yang, J.~Zhang,
  \href{https://doi.org/10.1109/ICDCS47774.2020.00032}{A collaborative learning
  framework via federated meta-learning}, in: 40th {IEEE} International
  Conference on Distributed Computing Systems, {ICDCS} 2020, Singapore,
  November 29 - December 1, 2020, {IEEE}, 2020, pp. 289--299.
\newblock \href {http://dx.doi.org/10.1109/ICDCS47774.2020.00032}
  {\path{doi:10.1109/ICDCS47774.2020.00032}}.
\newline\urlprefix\url{https://doi.org/10.1109/ICDCS47774.2020.00032}

\bibitem{VirginiaSmith_2017_NIPS_FLMTL}
V.~Smith, C.~Chiang, M.~Sanjabi, A.~S. Talwalkar,
  \href{https://proceedings.neurips.cc/paper/2017/hash/6211080fa89981f66b1a0c9d55c61d0f-Abstract.html}{Federated
  multi-task learning}, in: Advances in Neural Information Processing Systems
  30: Annual Conference on Neural Information Processing Systems 2017, December
  4-9, 2017, Long Beach, CA, {USA}, 2017, pp. 4424--4434.
\newline\urlprefix\url{https://proceedings.neurips.cc/paper/2017/hash/6211080fa89981f66b1a0c9d55c61d0f-Abstract.html}

\bibitem{PeterKairouz_2019_arXiv_FLProblems}
P.~Kairouz, H.~B. McMahan, B.~Avent, A.~Bellet, M.~Bennis, A.~N. Bhagoji,
  K.~Bonawitz, Z.~Charles, G.~Cormode, R.~Cummings, et~al., Advances and open
  problems in federated learning, arXiv preprint arXiv:1912.04977.

\bibitem{AshishVaswani_2017_NIPS_attention}
A.~Vaswani, N.~Shazeer, N.~Parmar, J.~Uszkoreit, L.~Jones, A.~N. Gomez,
  L.~Kaiser, I.~Polosukhin,
  \href{https://proceedings.neurips.cc/paper/2017/hash/3f5ee243547dee91fbd053c1c4a845aa-Abstract.html}{Attention
  is all you need}, in: Advances in Neural Information Processing Systems 30:
  Annual Conference on Neural Information Processing Systems 2017, December
  4-9, 2017, Long Beach, CA, {USA}, 2017, pp. 5998--6008.
\newline\urlprefix\url{https://proceedings.neurips.cc/paper/2017/hash/3f5ee243547dee91fbd053c1c4a845aa-Abstract.html}

\bibitem{AlanadeSantanaCorreia_2021_arXiv_attentionsurvey}
A.~de~Santana~Correia, E.~L. Colombini,
  \href{https://arxiv.org/abs/2103.16775}{Attention, please! {A} survey of
  neural attention models in deep learning}, arXiv preprint
  arXiv:2103.16775\href {http://arxiv.org/abs/2103.16775}
  {\path{arXiv:2103.16775}}.
\newline\urlprefix\url{https://arxiv.org/abs/2103.16775}

\bibitem{DavidWRomero_2021_ICLR_attentionvision}
D.~W. Romero, J.~Cordonnier,
  \href{https://openreview.net/forum?id=JkfYjnOEo6M}{Group equivariant
  stand-alone self-attention for vision}, in: 9th International Conference on
  Learning Representations, {ICLR} 2021, Virtual Event, Austria, May 3-7, 2021,
  OpenReview.net, 2021.
\newline\urlprefix\url{https://openreview.net/forum?id=JkfYjnOEo6M}

\bibitem{YaruHao_2021_AAAI_selfattention}
Y.~Hao, L.~Dong, F.~Wei, K.~Xu,
  \href{https://ojs.aaai.org/index.php/AAAI/article/view/17533}{Self-attention
  attribution: Interpreting information interactions inside transformer}, in:
  Thirty-Fifth {AAAI} Conference on Artificial Intelligence, {AAAI} 2021,
  Thirty-Third Conference on Innovative Applications of Artificial
  Intelligence, {IAAI} 2021, The Eleventh Symposium on Educational Advances in
  Artificial Intelligence, {EAAI} 2021, Virtual Event, February 2-9, 2021,
  {AAAI} Press, 2021, pp. 12963--12971.
\newline\urlprefix\url{https://ojs.aaai.org/index.php/AAAI/article/view/17533}

\bibitem{DimitriPBertsekas_2015_arXiv_Optimizationincrementaltype}
D.~P. Bertsekas, Incremental gradient, subgradient, and proximal methods for
  convex optimization: A survey, Optimization 2010~(2) (2015) 691--717.

\bibitem{lecun1998gradient}
Y.~LeCun, L.~Bottou, Y.~Bengio, P.~Haffner, Gradient-based learning applied to
  document recognition, Proceedings of the IEEE 86~(11) (1998) 2278--2324.

\bibitem{xiao2017fashion}
H.~Xiao, K.~Rasul, R.~Vollgraf,
  \href{http://arxiv.org/abs/1708.07747}{Fashion-mnist: a novel image dataset
  for benchmarking machine learning algorithms}, CoRR abs/1708.07747.
\newblock \href {http://arxiv.org/abs/1708.07747} {\path{arXiv:1708.07747}}.
\newline\urlprefix\url{http://arxiv.org/abs/1708.07747}

\bibitem{krizhevsky2009learning}
A.~Krizhevsky, G.~Hinton, et~al., Learning multiple layers of features from
  tiny images.

\end{thebibliography}

\end{document}